%% file: neurips_2025.tex
\documentclass{article}

    \PassOptionsToPackage{numbers, compress}{natbib}
 \usepackage[preprint]{neurips_2025}

\usepackage[utf8]{inputenc} 
\usepackage[T1]{fontenc}    
\usepackage{hyperref}       
\usepackage{url}            
\usepackage{booktabs}       
\usepackage{amsfonts}       
\usepackage{nicefrac}       
\usepackage{microtype}      
\usepackage{xcolor}         
\usepackage{natbib}

\usepackage{amsmath}
\usepackage{graphicx}
\usepackage{subcaption}
\usepackage{pgfplots}
\usepackage{multirow}
\usepackage{color,soul}
\usepackage{algorithm}
\usepackage{algpseudocode}
\usepackage{wrapfig}
\usepackage{textcomp}
\usepackage{tikz}
\usetikzlibrary{calc, fit, positioning, backgrounds, fadings, shapes.geometric, arrows.meta, decorations.pathreplacing}
\pgfplotsset{compat=1.18}
\definecolor{modernBlue}{RGB}{0, 114, 178}
\definecolor{modernGreen}{RGB}{0, 158, 115}
\definecolor{modernRed}{RGB}{190, 30, 45} 
\definecolor{modernPurple}{RGB}{112, 48, 160}

\title{InnerQ: Hardware-Aware Tuning-Free Quantization of KV Cache for Large Language Models}

\author{%
  Mohammadreza Tayaranian\thanks{Correspondence to \texttt{mohammadreza.tayaranian@mail.mcgill.ca}}  \quad Amir Ardakani \quad Warren J. Gross \\
  Department of Electrical and Computer Engineering\\
  McGill University\\
  Montreal, QC, Canada \\
}

\begin{document}

\maketitle

\begin{abstract}
When transformer-based language models are deployed for text generation, most of the inference time is spent in the decoding stage, where output tokens are generated sequentially.
Reducing the hardware cost of each decoding step is therefore critical for efficient long-context generation.
A major bottleneck is the key-value (KV) cache, whose size grows with sequence length and often dominates the model's memory footprint.
Prior work has proposed quantization methods to compress the KV cache while minimizing its loss of precision.
We present InnerQ, a hardware-aware KV cache quantization scheme that reduces decode latency without compromising evaluation performance.
InnerQ performs group-wise quantization by grouping cache matrices along their inner dimension.
This grouping strategy aligns dequantization with vector-matrix multiplication and increases data reuse across GPU compute units.
As a result, InnerQ reduces memory access and accelerates dequantization, achieving an average $1.3\times$ speedup over prior KV cache quantization methods and $2.7\times$ over the non-quantized baseline.
To maintain fidelity under aggressive compression, InnerQ incorporates three techniques: (i) hybrid quantization, which chooses symmetric or asymmetric quantization for each group based on local statistics; (ii) high-precision windows for both recent tokens and attention sink tokens to mitigate outlier leakage; and (iii) per-channel normalization of the key cache, computed once during prefill and folded into the model parameters to eliminate runtime overhead.
Beyond reducing latency, experiments on Llama and Mistral models show that InnerQ also improves few-shot evaluation scores relative to prior KV cache quantization methods.
\end{abstract}

\section{Introduction}

Large language models (LLMs) are increasingly deployed in text generation applications, including chatbots, many of which require processing long token sequences and therefore impose a growing demand on inference-time compute and memory resources.
Improving the efficiency of text generation is therefore essential for reducing power consumption and enabling deployment on resource-constrained edge devices.
A key contributor to generation cost is the decode phase, in which output tokens are produced autoregressively, one token at a time~\cite{leanattention}.

To improve decoding efficiency, modern LLMs commonly use KV caching~\cite{kvcache}, which stores the key and value matrices of previous tokens and restricts computation at each step to the newest token.
This mechanism reduces redundant computation and improves throughput.
However, because the cache grows linearly with the number of processed tokens, its memory footprint increases to the point where KV cache dominates the memory footprint of the model~\cite{minicache}.
The growing size and hardware footprint of the KV cache makes its compression a critical requirement for efficient long-sequence generation.

Previous work has proposed KV cache quantization as an effective compression strategy for reducing memory usage during decoding~\cite{kivi,gear,kvquant,squat}.
To better understand existing designs, consider the vector-matrix multiplication $C = AB$, where $A \in \mathbb{R}^{1 \times K}$ is a floating-point vector and $B \in \mathbb{R}^{K \times N}$ is a quantized matrix.
Depending on the operation, $A$ represents either the query or the attention weights, while $B$ represents either the key or the value matrix.
Existing methods apply group-wise quantization to $B$ by partitioning it into small groups, each with its own scale factor~\cite{kivi,squat}.
To preserve quantization accuracy, these groups are typically formed along the outer dimension, i.e., $N$, which helps isolate large outliers and reduce their effect on precision~\cite{kivi,squat}.

However, outer-dimension grouping is not well matched to the execution pattern of quantized vector-matrix multiplication on GPUs.
In this operation, each row of the cache matrix is first dequantized and then multiplied by the floating-point vector.
When groups are formed along the outer dimension, the entries within a row belong to different quantization groups and must be dequantized separately (Figure~\ref{fig:inner_outer:outer}).
In contrast, inner-dimension grouping allows compute units to reuse scale factors across multiple compute units, reducing memory access, improving data reuse, and ultimately lowering the latency of the quantized vector-matrix multiplication (Figure~\ref{fig:inner_outer:inner}).

\input{figure_inner_outer}
\input{figure_overview_quantization}

Building on this intuition, we propose InnerQ, a tuning-free, hardware-aware KV cache quantization method for efficient LLM decoding.
InnerQ is designed to reduce inference latency in text generation with quantized KV cache while preserving evaluation performance.
It provides three quantization variants, InnerQ\textsubscript{Base}, InnerQ\textsubscript{Hybrid}, and InnerQ\textsubscript{Small}, which trade off evaluation performance, cache size, and speedup.
To achieve this balance, we incorporate the following techniques into different InnerQ variants:
\begin{itemize}
    \item \textbf{Inner Dimension Grouping:}
    We apply group-wise quantization to the KV cache by forming groups along the \emph{inner} dimension of the cache matrices.
    As discussed earlier, this grouping strategy reduces memory access and speeds up dequantization.
    Experiments in Section~\ref{sec:speedup} demonstrate that inner-dimension grouping, adopted by all InnerQ variants, reduces decoding latency compared to KIVI~\cite{kivi}, which groups KV cache matrices along the outer dimension.
    This highlights inner-dimension grouping as a key design choice for efficient quantized KV caching.
    \item \textbf{High-precision Windows:}
    Inspired by prior work~\cite{kivi,squat,skvq,streamingllm}, InnerQ retains a small subset of cache tokens in half-precision.
    Unlike KIVI~\cite{kivi}, which allocates the entire high-precision window to the most recent tokens, InnerQ reserves part of this budget for the first few tokens in the sequence.
    These initial tokens often act as \emph{attention sinks}, receiving disproportionately high attention scores throughout generation.
    This design preserves the attention sink tokens and prevents their influence from leaking into the quantization groups and diminishing the precision of other tokens~\cite{attention_sink,streamingllm}.
    \item \textbf{Per-channel Normalization of Key Cache:}
    To reduce the magnitude of outliers, InnerQ applies channel-wise normalization to the key cache.
    The normalization factors are computed during the prefill stage and folded into the key and query weights, thereby avoiding any additional runtime overhead.
    Token-wise grouping of the key cache makes per-channel normalization straightforward, whereas outer-dimension grouping makes it less natural and more costly to integrate.
    \item \textbf{Hybrid Quantization:}
    Symmetric and asymmetric quantization are the most prominent quantization methods used in the KV cache.
    In short, symmetric quantization centers the data around zero while asymmetric quantization shifts the data to the positive range before quantization.
    However, as discussed in Section~\ref{sec:ablation_q_mode}, we observe that, given their inherent assumptions about the underlying data distribution, these quantization modes sometimes fail to maintain KV cache precision.
    Notably, the value cache suffers from precision loss when quantized using asymmetric quantization in low-bit regimes.
    To avoid a costly hyper-parameter search, we propose a hybrid quantization scheme that allows each quantization group in the value cache to dynamically select either symmetric or asymmetric quantization based on the values in that group.
    Our experiments show that, compared to non-adaptive quantization modes, hybrid quantization improves the evaluation score of quantization in low-bit regimes (Section~\ref{sec:accuracy}) with a negligible effect on latency (Section~\ref{sec:speedup}).
\end{itemize}

We evaluate InnerQ on Llama~\cite{llama2,llama3} and Mistral~\cite{mistral} models of varying sizes and measure the resulting evaluation performance under KV cache quantization.
As detailed in Section~\ref{sec:accuracy}, InnerQ achieves average evaluation scores comparable to those of the non-quantized KV cache and prior KV cache quantization methods.
We also implement fused CUDA kernels for quantized KV cache operations and measure their latency on an NVIDIA Jetson edge platform.
Our measurements, discussed in Section~\ref{sec:speedup}, show that InnerQ achieves an average speedup of $2.7\times$ relative to a non-quantized \texttt{FP16} cache.
We also compare the latency of InnerQ with prominent KV cache quantization methods KIVI~\cite{kivi} and TurboQuant~\cite{turboquant} and observe an average speedup of $1.3\times$ and $1.2\times$, respectively.

\section{Related Work}

KIVI \cite{kivi} is one of the earliest works that proposes a tuning-free KV cache quantization method.
It cites channel outliers as the main challenge in KV cache quantization and proposes an asymmetric quantization over the \textit{outer} dimension of the key and value matrices to address this challenge.
While we also propose a tuning-free quantization method with minimal overhead, we focus on quantizing the KV cache over the \emph{inner} dimension to achieve higher speed up.

TurboQuant is a recent tuning-free, data-oblivious vector quantization method for high-dimensional vectors, including KV cache representations~\cite{turboquant}. 
It applies a random rotation to the input vectors so that the rotated coordinates follow a concentrated Beta distribution, and then quantizes these coordinates using optimal non-uniform scalar quantizers~\cite{turboquant}. 
The corresponding codebooks are obtained by solving an optimization problem for each target bit-width and can be precomputed, avoiding data-dependent calibration. 
This design differs from InnerQ, which applies uniform group-wise quantization to the KV cache; TurboQuant instead relies on non-uniform coordinate-wise quantization after random rotation, with an effective KV cache precision of $3.5$ bits, obtained by assigning different bit-widths to outlier and non-outlier channel sets. 
Based on our experimental results, InnerQ achieves a comparable evaluation score to TurboQuant (Section~\ref{sec:accuracy}) while having a lower latency (Section~\ref{sec:speedup}).

GEAR \cite{gear} uses low-rank decomposition alongside quantization to address the outlier challenge.
Despite its low accuracy drop, its need to solve small optimization problems for the low-rank decomposition leads to runtime overhead.
It proposes a CUDA implementation to further mitigate this overhead.
SQuat \cite{squat} proposes to minimize the quantization error by quantizing the key matrix such that its quantization error is orthogonal to the query.
Building on KIVI, SQuat requires solving an optimization problem periodically to find the orthogonal matrix.
Similar to GEAR, this extra optimization leads to improved accuracy, while negatively affecting the throughput of the quantized attention.
Unlike GEAR and SQuat, our proposed method does not add extra steps to the quantization and is shown to have negligible overhead compared to KIVI.

KVQuant \cite{kvquant} proposes a non-uniform quantization where the quantization signposts are learned for each tensor.
It uses a calibration dataset to compute the fisher matrix and find the signposts before inference begins.
SKVQ \cite{skvq} uses per-token group-wise quantization with dynamic clipping for both key and value.
In other words, while the key cache is quantized over the inner dimension, the value cache is quantized over the outer dimension.
Similar to our work, they propose to keep the most recent tokens in floating-point for higher precision.
To overcome outlier channels, SKVQ proposes a custom ordering of channels.
The optimal ordering of channels is found by gathering features of channels on a calibration dataset and clustering similar channels together for a smoother quantization.
They also propose a variable group size which enables them to group similar channels together.
However, whether their variable group size efficiently uses resources on a general-purpose GPU is not demonstrated by the authors.
While calibration is an important step for both SKVQ and KVQuant, it leads to extra overhead and also makes their method reliant on external data.
Unlike these methods, our proposed method dynamically adapts to data with hybrid quantization (Section \ref{sec:hybrid_quantization}) and does not require any extra calibration.

\section{Background}

\subsection{General Notation}
\label{sec:notation}

Let $A \in \mathbb{R}^{M \times K}$ be a matrix.  
We adopt a slicing notation analogous to that used in array-based libraries like PyTorch, where the $i$-th row and the $j$-th column of $A$ are denoted by $A_{i,:}$ and $A_{:,j}$, respectively.
For indices $i_1 \leq i_2$, we denote by $A_{i_1:i_2,:}$ the submatrix consisting of rows $i_1$ through $i_2$ (inclusive).
Similarly, for indices $j_1 \leq j_2$, the submatrix consisting of columns $j_1$ through $j_2$ (inclusive) is denoted by $A_{:,j_1:j_2}$.
The notation $A_{i_1:,:}$ refers to the submatrix formed by rows $i_1$ through $M$, while $A_{:i_1,:}$ denotes the submatrix consisting of rows $1$ through $i_1$, both inclusive.
Given matrices $A \in \mathbb{R}^{N \times K}$ and $B \in \mathbb{R}^{M \times K}$, their vertical concatenation (along the first dimension) is denoted by
$[A; B] \in \mathbb{R}^{(N+M) \times K}$.
This notation extends naturally to higher-dimensional arrays, where slicing along each axis follows the same convention.

\subsection{KV Caching}

Multi-head attention (MHA) is one of the building blocks of transformer-based large language models.
It starts by transforming the hidden states $h$ into $Q \in \mathbb{R}^{n_h \times N_Q \times d_h}$ and to $K$ and $V \in \mathbb{R}^{n_h \times N_K \times d_h}$ as
\begin{equation}
\label{eq:att:q}
    Q=hW_Q
\end{equation}
\begin{equation}
\label{eq:att:kv}
    K=hW_K
    \quad \quad
    V=hW_V
\end{equation}
where $n_h$ is the number of attention heads and $d_h$ is the head dimension.
It then computes the attention matrix, $O \in \mathbb{R}^{N_Q \times d}$, as
\begin{equation}
\label{eq:att:1}
    S=QK^{\top}
\end{equation}
\begin{equation}
\label{eq:att:2}
    P=\mathrm{softmax} \left( \frac{S}{\sqrt{d_h}} \right)
\end{equation}
\begin{equation}
\label{eq:att:3}
    O=PV
\end{equation}
where $d=n_h \times d_h$ is the hidden dimension of the model.

When a language model is used for language generation its operation, is divided into two stages of \textit{prefill} and \textit{decode}.
During the prefill stage, the hidden states $h \in \mathbb{R}^{N_Q \times d}$ contain multiple tokens that help the model obtain the context of the current task.
After prefill, the model enters the decode stage where it generates tokens one by one ($N_Q = 1$).
Despite that, key and value matrices contain all the context from the prefill stage and the decode stage up to that point ($N_K > 1$).

Since these matrices grow with every new token, KV caching proposes to cache and reuse key and value matrices to avoid their recomputation for each step.
KV cache matrices are initialized at the end of the prefill stage as
\begin{equation}
\label{eq:att:kv_cache_init}
    K_\mathrm{cache}=K \quad\quad V_\mathrm{cache}=V
\end{equation}
and are re-used for the subsequently generated tokens so that at each step of the decode phase the key and value are computed only for the last token.
Using KV caching, Equation \ref{eq:att:kv} is re-written as
\begin{equation}
\label{eq:att:kv_cached}
    K=[K_\mathrm{cache};hW_K]
    \quad\quad
    V=[V_\mathrm{cache};hW_V].
\end{equation}
After each decode step, the cache is updated with the most recently computed key and value.
This technique speeds up the generation process by avoiding the full linear transformation that creates the full $K$ and $V$ for each token.

\subsection{KV Cache Quantization}
\label{sec:kv_cache_quantization}

Previous work has proposed quantizing the KV cache to reduce its memory footprint.
Given that both $K$ and $V$ are multiplied by full-precision matrices, KV cache quantization methods store them in low-precision data types, but dequantize them before the matrix multiplications of Equations \eqref{eq:att:1} and \eqref{eq:att:3}.
Applying quantization to the KV cache, Equation \eqref{eq:att:kv_cache_init} is reformulated as
\begin{equation}
\label{eq:att:kv_cache_init_quantized}
    \hat{K}_\mathrm{cache}=\mathrm{quant}(K) \quad\quad \hat{V}_\mathrm{cache}=\mathrm{quant}(V)
\end{equation}
and Equation \eqref{eq:att:kv_cached} is replaced by
\begin{equation}
\label{eq:att:kv_cached_quantized}
    K=[\mathrm{dequant}(\hat{K}_\mathrm{cache});hW_K]
    \quad \quad
    V=[\mathrm{dequant}(\hat{V}_\mathrm{cache});hW_V].
\end{equation}
where $\hat{K}_\mathrm{cache}$ and $\hat{V}_\mathrm{cache}$ are quantized cache matrices.

\section{Methodology}

\subsection{Quantization Scheme}
\label{sec:quantization}

We propose to use $b$-bit group-wise quantization to quantize $K,V \in \mathbb{R}^{n_h \times N_K \times d_h}$ with a group size $G$.
Each matrix is divided into smaller quantization groups where each group has a scale factor and zero-point (if applicable).

We begin by describing the symmetric and asymmetric quantization modes and then describe our proposed hybrid quantization mode.
Although $K$ and $V$ are interchangeable in the equations and definitions, we use $K$ throughout for ease of reading.
Without loss of generality we assume that $K$ is grouped over the inner dimension ($d_h \equiv 0 \pmod{G}$).

\subsubsection{Asymmetric and Symmetric Quantization}
\label{sec:asym_sym_quantization}

Group-wise asymmetric quantization represents each tensor using a scale factor and a zero-point.
Let $S \in \mathbb{R}^{n_h \times N_K \times \frac{d_h}{G}}$ and $Z \in \mathbb{R}^{n_h \times N_K \times \frac{d_h}{G}}$ denote the corresponding scale factor and zero-point matrices, respectively.
The quantized value of the element for head $i$, token $j$, and channel $k$ of $K$ is then given by

\begin{equation}
\label{eq:asym_quant}
    \hat{K}_{i,j,k}=\mathrm{quant}_{\mathrm{asym}}(K_{i,j,k}) = \mathrm{clip}\left(\left\lfloor\frac{K_{i,j,k}-Z_{i,j,g}}{S_{i,j,g}}\right\rceil, 0, 2^b -1\right)
\end{equation}
where $\lfloor \cdot \rceil$ is the round-to-nearest function and $g = \left\lfloor \frac{k-1}{G} \right\rfloor + 1$ is the group index of $K_{i,j,k}$.
The corresponding zero point is computed as $Z_{i,j,g}=\min(\mathcal{G}_{i,j,g})$ and its scale factor as
\begin{equation}
    S_{i,j,g}=\frac{\mathrm{max}(\mathcal{G}_{i,j,g})-Z_{i,j,g}}{2^b-1}
\end{equation}
where $\mathcal{G}_{i,j,g}=\{K_{i,j,k}|\lfloor(k-1)/G\rfloor=g\}$ is the set of values in the group.
As the values are shifted by their minimum, for all $i$, $j$, and $k$ we have $0\leq\hat{K}_{i,j,k}$ and thus $\hat{K}$ is represented with an unsigned data type.
Similarly, $\hat{K}_{i,j,k}$ is dequantized as
\begin{equation}
\label{eq:asym_dequant}
    \mathrm{dequant}_\mathrm{asym}(\hat{K}_{i,j,k})=S_{i,j,g}\hat{K}_{i,j,k}+Z_{i,j,g}.
\end{equation}

Symmetric quantization is equivalent to asymmetric quantization, except that the zero-point is fixed to $0$ for all groups.
Accordingly, the quantization and dequantization processes follow Equations~\eqref{eq:asym_quant} and~\eqref{eq:asym_dequant} with $Z_{i,j,g}=0$.
The scale factor for each group is then computed using the maximum absolute value as
\begin{equation}
    S_{i,j,g}=\frac{\mathrm{max}(|\mathcal{G}_{i,j,g}|)}{2^b-1}
\end{equation}
where $|\cdot|$ is the absolute value operator.
Here, unlike asymmetric quantization, quantized values can be either negative or positive and are represented using signed data types.

\subsubsection{Hybrid Quantization}
\label{sec:hybrid_quantization}

\input{figure_hybrid}

As their names suggest, symmetric and asymmetric quantization impose different assumptions on the data distribution. 
The extent to which these assumptions are satisfied directly influences the fidelity of the quantized representation. 
To illustrate, consider a quantization group $\mathcal{G}$ such that $\min(\mathcal{G}) > 0$. 
Symmetric quantization allocates a sign bit to represent the quantized values in $\mathcal{G}$.
For an entirely positive group, this reduces effective resolution by reserving part of the range for values that never occur.
In contrast, asymmetric quantization exploits the nonnegativity of $\mathcal{G}$ by shifting the quantization interval, thereby using the available bit budget more efficiently to represent magnitudes rather than an unused negative range.

Empirically, the distribution of activation and weight values varies substantially across models and layers (see Figure~2 in~\cite{kivi}). 
This variation suggests that the preferable choice between symmetric and asymmetric quantization is configuration-dependent rather than universal. 
Consistent with this observation, our ablation results in Section~\ref{sec:ablation_q_mode} show that model performance can be sensitive to the quantization mode.

To address this sensitivity, we propose a \emph{hybrid quantization} scheme in which each quantization group independently selects between symmetric and asymmetric quantization.
Following the notation of Section~\ref{sec:asym_sym_quantization}, we introduce a binary mask $\mathcal{M} \in \{0,1\}^{n_h \times N_K \times \frac{d_h}{G}}$, where $\mathcal{M}_{i,j,g}=1$ indicates that $\mathcal{G}_{i,j,g}$ is quantized using the asymmetric mode, and $\mathcal{M}_{i,j,g}=0$ indicates that it is quantized using the symmetric mode.
As depicted in Figure \ref{fig:hybrid}, we determine $\mathcal{M}_{i,j,g}$ by finding the quantization mode with the lower reconstruction error.

Following the notation of Section~\ref{sec:asym_sym_quantization}, we have
\begin{equation}
\label{eq:hybrid_dequant}
    \mathrm{dequant}_{\mathrm{hybrid}}(\hat{K}_{i,j,k})
    = S_{i,j,g}\hat{K}_{i,j,k} + \mathcal{M}_{i,j,g} Z_{i,j,g}.
\end{equation}
When $\mathcal{M}_{i,j,g}=0$, the zero-point term does not contribute to dequantization.
Although our experiments reveal that hybrid quantization rarely uses asymmetric quantization for groups, we store the zero-points as a dense matrix to avoid the latency overhead associated with sparse representations.
We study the sparsity of $\mathcal{M}$ and its impact on the latency of dequantization operations in Section~\ref{sec:ablation_sparsity_m}.
Since scale factors are strictly positive, we repurpose their sign bit to encode $\mathcal{M}_{i,j,g}$.
We further discuss the hardware-aware implementation of the hybrid quantization mode in Section~\ref{sec:speedup}.

\subsection{High-precision Window}
\label{sec:fp_window}

As shown in previous work, keeping a small window of KV cache tokens in a high-precision data type greatly helps maintain model accuracy \cite{kivi,skvq,streamingllm,squat}.
We keep the last $w_\mathrm{recent}$ tokens in half-precision and denote them as $K_{\mathrm{recent}}$ and $V_{\mathrm{recent}}$.
Similarly, the first $w_\mathrm{sink}$ tokens are also kept in half-precision and are referred to as $K_{\mathrm{sink}}$ and $V_{\mathrm{sink}}$.
Given that finding the exact tokens that act as attention sinks appear incurs additional computational overhead~\cite{attention_sink,streamingllm}, we set a fixed $w_\mathrm{sink}$ for all setups.
We study the effect of changing $w_\mathrm{sink}$ and $w_\mathrm{recent}$ in Section~\ref{sec:ablation_w_size}.

Incorporating the high-precision windows into Equation \eqref{eq:att:kv_cache_init_quantized}, we have
\begin{equation}
\begin{aligned}
\label{eq:att:kv_cache_init_quantized_window}
K_\mathrm{sink}=K_{:,:w_\mathrm{sink},:} \quad & \quad V_\mathrm{sink}=V_{:,:w_\mathrm{sink},:} \\
K_\mathrm{recent}=K_{:,-w_\mathrm{recent}:,:} \quad & \quad V_\mathrm{recent}=V_{:,-w_\mathrm{recent}:,:}\\
\hat{K}_\mathrm{cache}=\mathrm{quant}(K_{:,w_\mathrm{sink}:-w_\mathrm{recent},:}) \quad & \quad \hat{V}_\mathrm{cache}=\mathrm{quant}(V_{:,w_\mathrm{sink}:-w_\mathrm{recent},:}).
\end{aligned}
\end{equation}

During the decode phase, $K_\mathrm{sink}$ and $V_\mathrm{sink}$ remain fixed, while the newly generated $K$ and $V$ for each token are appended to $K_\mathrm{recent}$ and $V_\mathrm{recent}$.
As tokens accumulate in $K_\mathrm{recent}$ and $V_\mathrm{recent}$, the oldest ones are quantized and moved to $\hat{K}_\mathrm{cache}$ and $\hat{V}_\mathrm{cache}$.
While in per-channel group-wise quantization the old tokens can be quantized one by one, per-token group-wise quantization requires the number of quantized tokens to be a multiple of the group size.

\subsection{Per-channel Normalization of $K$}
\label{sec:normalization}

Outliers within a quantization group reduce the effective precision available to the remaining values.
Prior work has shown that such outliers emerge along the channel dimension of $K$ \cite{kivi}.
To mitigate this effect, we apply per-channel normalization to $K$ which is beneficial in mitigating outliers when quantization groups span multiple channels, as in per-token group-wise quantization (our work).
Specifically, the normalization factor for channel $k$ is defined as $\mathrm{norm}^K_k = \sqrt{\max(|K_{:,:,k}|)}$, where $|\cdot|$ denotes the absolute value.
The vector $\mathrm{norm}^K$ is computed once at the end of the prefill stage and absorbed into $W_Q$ and $W_K$, thereby hiding the normalization overhead during decoding.

\subsection{InnerQ: Quantizing Key and Value Over the Inner Dimension}
\label{sec:innerq}

At each step of the decode phase, quantized KV cache matrices are dequantized and then used by Equations \eqref{eq:att:1} and \eqref{eq:att:3} to compute the output.
Since $N_Q=1$ in the decode stage, Equations \eqref{eq:att:1} and \eqref{eq:att:3} are implemented as vector-matrix multiplication operations (GEMV).
Given the high latency of memory operations in modern GPU hardware, it is suboptimal to perform the dequantization and GEMV operations separately.
Instead, as suggested by previous work \cite{kivi}, these operations are performed in a fused kernel where each row of the quantized matrix is dequantized and then multiplied by the floating-point vector.
The fused kernel allows the output of the dequantization operation to directly pass to the GEMV operation without using the high-latency global memory to store the intermediate results.

We propose to align the quantization groups with the dimension over which the multiplication is performed, i.e. the \emph{inner} dimension.
As depicted in Figure \ref{fig:inner_outer:inner}, when grouped over the inner dimension, values of the same group in each channel of $K$ have the same scale factor\footnote{A similar argument is valid for zero-points in the case of asymmetric quantization.}.
This enables compute units processing the same channel group to load the scale factor once and reuse it.
In contrast, outer dimension quantization, where grouping is done over the tokens, loads multiple scale factors per row.
In this scheme, compute units do not reuse these values and each loads its own scale factor which leads to multiple repeated high-bandwidth memory loads. 
As a result of the aligned memory access patterns, inner grouping can achieve a higher throughput and improve the overall latency of the attention operation.
The effect of aligned memory access is even more pronounced in edge devices where memory bandwidth and cache size are limited.
A similar argument is valid for Equation \eqref{eq:att:3} where $P$ is a floating-point vector and $V$ is a quantized matrix in which the token dimension is the inner dimension and the channel dimension is the outer dimension.
In Section~\ref{sec:speedup}, we validate this hypothesis by comparing the latency of quantized GEMV kernels under inner- and outer-dimension grouping.

We present \textbf{InnerQ}, a hardware-aware KV cache quantization method designed to reduce the latency of large language model inference during the decode phase.
We introduce three variants of InnerQ to allow a tunable trade-off between numerical precision, inference latency, and memory footprint.
The variants differ primarily in how they quantize $V_\textrm{cache}$, while sharing the same strategy for $K_\textrm{cache}$.
We first describe the quantization scheme shared by all variants and then summarize the variant-specific choices for $V_\textrm{cache}$ quantization.

All variants apply inner dimension grouping to both key and value matrices.
Specifically, $K_\textrm{cache}$ is quantized using per-token group-wise quantization, and $V_\textrm{cache}$ is quantized using per-channel group-wise quantization.
For $K_\textrm{cache}$, we use 3-bit symmetric quantization in all variants; thus each $\hat{K}_{i,j,k}$ is represented as a 3-bit signed integer.
Consistent with Section~\ref{sec:fp_window}, we maintain high-precision windows for recent and sink tokens.
We also apply per-channel normalization to $K$ following Section~\ref{sec:normalization} to reduce sensitivity to outliers.

\textbf{InnerQ\textsubscript{Base}} quantizes $V_\textrm{cache}$ using 3-bit symmetric quantization.
Based on the observation that $V_\textrm{cache}$ has a higher tolerance for quantization \cite{quantizewhatcounts}, \textbf{InnerQ\textsubscript{Small}} applies 2-bit symmetric quantization to $V_\textrm{cache}$ and reduces bandwidth and latency, at the expense of reduced dynamic range.
To compensate for this limitation, \textbf{InnerQ\textsubscript{Hybrid}} adopts 2-bit hybrid quantization (Section \ref{sec:hybrid_quantization}) for $V_\textrm{cache}$, achieving higher fidelity at a modest storage and runtime cost relative to InnerQ\textsubscript{Small}.

\section{Empirical Evaluation}
\label{sec:experiments}

\subsection{Evaluation Metric Score}
\label{sec:accuracy}

We apply InnerQ to Llama \cite{llama2,llama3} and Mistral \cite{mistral} language models and observe its effect on the metric score of the model.
We implement a simulated quantized KV cache based on the Hugging Face Transformers library \cite{huggingface} and use the LM-eval \cite{lmeval} framework to build our evaluation script.
We compare our results against non-quantized \texttt{FP16} KV cache as the baseline.
We also compare our work with KIVI \cite{kivi} as a tuning-free KV cache quantization method that uses group-wise quantization.
For a better comparison, we provide an additional setup denoted as KIVI\textsubscript{Sink} by following Section~\ref{sec:fp_window} and devoting a portion of the high-precision window to the sink tokens.
Following previous work we set the group size to $32$ and the total length of the high-precision window to $128$.
For KIVI\textsubscript{Sink} and InnerQ we have $w_\mathrm{sink}=32$ and $w_\mathrm{recent}=96$ and for KIVI we have $w_\mathrm{sink}=0$ and $w_\mathrm{recent}=128$.
In addition to KIVI, we run evaluation experiments while quantizing the KV cache with the recently proposed TurboQuant method~\cite{turboquant}.
\input{table_results_main}
As our experiments and models use softmax-based attention, we use the MSE variant of TurboQuant with a high-precision window size of $128$.
We also set the key bit-width and value bit-width of TurboQuant to $4$ and $3$, respectively, which makes it comparable in size to InnerQ\textsubscript{Base} (Table~\ref{tab:bits}).
For KIVI and InnerQ we use our own quantized KV cache implementation while for TurboQuant we use a community implementation \cite{turboquant_code}.

We compare the performance of InnerQ with other setups by applying them to pre-trained models and running few-shot evaluation on the evaluation set of mathematical reasoning tasks GSM8K \cite{gsm8k} and Minerva Math \cite{minerva}.
We also use the instruction-tuned variants of the language models and run evaluation experiments on few-shot code-generation tasks HumanEval \cite{humaneval} and MBPP \cite{mbpp}.

Our experimental results in Table~\ref{tab:accuracy} identify InnerQ\textsubscript{Base} as the most accurate variant of our method.
On average, it improves the score of KIVI and KIVI\textsubscript{Sink} by $1.8$ and $3.5$ points, respectively.
InnerQ\textsubscript{Base} also achieves an average gain of $0.4$ points over TurboQuant.
A more fine-grained analysis shows that InnerQ\textsubscript{Base} has a higher average score than TurboQuant on GSM8K, HumanEval, and MBPP tasks.

Among the InnerQ variants, InnerQ\textsubscript{Small} exhibits an average degradation of $3$ points relative to InnerQ\textsubscript{Base}, which can be attributed to its 2-bit value cache.
In contrast, the hybrid quantization strategy used in InnerQ\textsubscript{Hybrid} reduces the average gap to InnerQ\textsubscript{Base} to $1.7$ points, while achieving an average score comparable to KIVI\textsubscript{Sink}.

A secondary observation from Table~\ref{tab:accuracy} is the performance gap between KIVI and KIVI\textsubscript{Sink}. This gap indicates that introducing a high-precision sink window into KIVI has a substantial effect on performance, increasing the average score of KIVI\textsubscript{Sink} by $1.7$ points over KIVI.
We investigate the effect of the high-precision window size in Section~\ref{sec:ablation_w_size}.

\input{table_results_longbench}

We further evaluate the proposed KV cache quantization methods in long-sequence generation settings using selected tasks from the LongBench benchmark~\cite{longbench}.
The results, reported in Table~\ref{tab:score_longbench}, indicate that the benefit of the high-precision sink window diminishes as the sequence length increases.
In several cases, KIVI outperforms KIVI\textsubscript{Sink}; on average, KIVI\textsubscript{Sink} improves over KIVI by only $0.2$ points.
InnerQ\textsubscript{Base} remains competitive with existing KV cache quantization methods, improving the average score over KIVI and KIVI\textsubscript{Sink} by $0.4$ and $0.2$ points, respectively.

Consistent with the results in Table~\ref{tab:accuracy}, InnerQ\textsubscript{Small} incurs an average degradation of $4.3$ points relative to InnerQ\textsubscript{Base}, further highlighting the sensitivity of model performance to aggressive value-cache quantization.
Similar to results on the shorter-context benchmarks, InnerQ\textsubscript{Hybrid} recovers a substantial portion of the performance loss introduced by low-bit value-cache quantization.
On average, InnerQ\textsubscript{Hybrid} lags behind InnerQ\textsubscript{Base} by only $0.5$ points while outperforming InnerQ\textsubscript{Small} by $3.8$ points.
The performance of InnerQ\textsubscript{Hybrid} across the tasks reported in Tables~\ref{tab:accuracy} and~\ref{tab:score_longbench} demonstrates the effectiveness of hybrid quantization for adapting value-cache quantization in low-bit regimes.

Overall, the experimental results indicate that InnerQ\textsubscript{Base} matches or surpasses existing KV cache quantization methods in evaluation metric score.
As shown in Section~\ref{sec:speedup}, InnerQ\textsubscript{Base} also reduces the latency associated with using the quantized KV cache relative to both KIVI and TurboQuant.
Taken together, the metric and latency results further demonstrate the effectiveness of InnerQ\textsubscript{Hybrid}, which preserves comparable metric score while maintaining low latency.

\subsection{Bit-width}
\label{sec:bitwidth}

\input{table_bits}

Table \ref{tab:bits} summarizes the total effective bit-width of various KV cache quantization methods, including both the base integer bit-width and the overhead of scale factors, zero-points, and channel norms.
With a group size of $32$ for KIVI and InnerQ, storing an \texttt{FP16} scale factor (zero-point) contributes $\frac{16}{32}=0.5$ additional bits per quantized number.
Although the zero-point matrix in InnerQ\textsubscript{Hybrid} is highly sparse, we still budget $0.5$ bits for the zero-point overhead and defer exploiting sparsity to future work.
To measure the overhead of storing channel norms for TurboQuant, we assume a head dimension of $128$ and \texttt{FP32} channel norms, which leads to $0.25$ bits of overhead per quantized number.

\subsection{Latency}
\label{sec:speedup}

\input{table_latency_dequant}
\input{graph_speedup_figure}

We evaluate how different KV cache quantization methods affect the latency of token generation in the decode phase.
To this end, we implement fused CUDA kernels that perform dequantization and GEMV on their input.
The fused kernel takes a quantized matrix and an \texttt{FP16} vector to simulate Equations \eqref{eq:att:1} and \eqref{eq:att:3} in the decode phase with a quantized KV cache.
We set the batch size to $1$ to reflect the interactive usage of an end-user generating text on an edge device.
We run our latency measurements on the NVIDIA Jetson Xavier NX embedded platform.
Reported latencies (in microseconds) are collected after $10$ warm-up iterations and are averaged over $100$ measured runs.
Table \ref{tab:latencies_dequant} reports the latency of Equations \eqref{eq:att:1} and \eqref{eq:att:3} using quantized key and value caches for one layer of Llama 3.2-8B across different sequence lengths.
We assume that for the value cache in InnerQ\textsubscript{Hybrid} $\mathcal{M}$ is $99\%$ sparse.
Figure \ref{fig:speedups} summarizes these measurements by plotting the speedup of each InnerQ variant over the \texttt{FP16} baseline, KIVI, and TurboQuant.

As discussed in Section~\ref{sec:innerq}, the proposed InnerQ variants differ mainly in their quantization policy for the value cache.
Our experiments show that InnerQ\textsubscript{Small} achieves the lowest latency due to its lower bit-width.
InnerQ\textsubscript{Hybrid} is slightly slower due to its higher computational demand and the need to load additional zero-points for some groups.
We further study the impact of the level of sparsity of $\mathcal{M}$ in the hybrid quantization scheme on the latency of the fused kernel in Section~\ref{sec:ablation_sparsity_m}.
As shown in the leftmost part of Figure~\ref{fig:speedups}, all InnerQ variants achieve a comparable average speedup of $2.7 \times$ relative to the \texttt{FP16} baseline which steadily rises as the sequence length grows.

Compared with KIVI, all InnerQ variants are notably faster and achieve consistently lower latency across different sequence lengths.
The speedup of InnerQ\textsubscript{Base} and InnerQ\textsubscript{Hybrid} relative to KIVI exceeds $1.2\times$ despite their average bit-widths being higher than KIVI.
This highlights the importance of data reuse and its potential to reduce the latency of quantized KV cache operations.
As the sequence length grows, the latency gap between KIVI and InnerQ variants widens in favor of InnerQ.
Notably, when $32768$ tokens are stored in the KV cache, the relative speedup of InnerQ\textsubscript{Small} compared to KIVI is nearly $1.4 \times$.

TurboQuant employs a precomputed codebook for dequantization, along with online per-channel normalization to rescale the dequantized key and value matrices to their original range.
We omit the normalization step for the value cache and assume that its contribution to the latency is negligible as it can be fused into the softmax operation.
Although TurboQuant requires fewer normalization factors than the scaling factors used in InnerQ, the codebook lookup requires multiple accesses to CUDA shared memory, thereby increasing kernel latency.
As illustrated in Figure~\ref{fig:speedups}, \textsc{InnerQ}\textsubscript{Base} achieves a relatively consistent $1.2\times$ speedup over TurboQuant.
Other InnerQ variants further improve the speedup to approximately $1.3\times$, at the expense of a reduction in evaluation performance as detailed in Section~\ref{sec:accuracy}.

Beyond the fused dequantization and GEMV kernels, we also develop CUDA kernels to handle quantization for each method.
As noted in Section~\ref{sec:fp_window}, tokens evicted from the high-precision windows $K_\textrm{recent}$ and $V_\textrm{recent}$ are quantized and stored in $\hat{K}_\mathrm{cache}$ and $\hat{V}_\mathrm{cache}$.
The eviction pattern is governed by the dimension across which quantization groups are defined in the cache.
InnerQ employs per-token group-wise quantization for the key cache and per-channel group-wise quantization for the value cache, whereas KIVI uses per-channel quantization for keys and per-token quantization for values.
This design leads InnerQ to quantize one key token at every step, while value tokens are evicted and quantized in groups of $G$ ($32$) every $32$ steps.
Conversely, KIVI evicts and quantizes $32$ key tokens every $32$ steps and one value token at each step.
Unlike KIVI and InnerQ, which are group-wise quantization methods, TurboQuant quantizes one key and one value token at each step.

\input{table_latency_quant}

Based on their eviction behavior, we measure and report the average quantization latency overhead during the decode phase.
As shown in Table~\ref{tab:latencies_quant}, the latency gap between KIVI and InnerQ is marginal, amounting to approximately $2$~\textmu s.
Because TurboQuant performs quantization more frequently, it incurs a higher quantization overhead.
Considering the latencies reported in Table~\ref{tab:latencies_dequant}, this additional $14$~\textmu s overhead can reduce the overall throughput of TurboQuant, particularly for small- to medium-length sequences.
Unlike dequantization, which lies on the critical path for producing the attention output, quantization of newly evicted tokens from $K_\textrm{recent}$ and $V_\textrm{recent}$ does not directly impact output generation.
This property provides an opportunity to pipeline the quantization process during periods of GPU idleness and mask its latency.
We leave the design and evaluation of such pipelining strategies to future work.

\section{Ablation Studies}

\subsection{High-precision Window Size}
\label{sec:ablation_w_size}

\input{graph_ablation_figure}

To analyze the effect of high-precision windows on models with quantized KV caches, we vary the hyperparameters $w_\mathrm{sink}$ and $w_\mathrm{recent}$.
We maintain a fixed number of total high-precision tokens by adjusting $w_\mathrm{sink}$ and setting $w_\mathrm{recent} = 128 - w_\mathrm{sink}$.

Figure~\ref{fig:ablation:sink} illustrates the evaluation performance of different models across window configurations and tasks.
Our findings reveal that the impact of window size on performance is highly dependent on both the model and the task.
Nonetheless, in many cases, allocating even a modest high-precision window of size $32$ yields a substantial accuracy improvement.
This behavior is especially pronounced for KIVI, InnerQ\textsubscript{Hybrid}, and InnerQ\textsubscript{Small}, whereas InnerQ\textsubscript{Base} demonstrates stable performance even when no high-precision sink window is used.

We also observe that in certain scenarios, such as GSM8K on Mistral 7B v0.3, further increasing $w_\mathrm{sink}$ at the expense of $w_\mathrm{recent}$ has a negligible effect on the metric score.
In contrast, other scenarios show a clear performance optimum, underscoring the importance of balancing the sizes of the two windows.
We defer a systematic investigation of optimal window sizes and the selection of tokens included in each window to future work.

\subsection{Sparsity of $\mathcal{M}$}
\label{sec:ablation_sparsity_m}

\input{table_ablation_hybrid}

Section~\ref{sec:hybrid_quantization} defines $\mathcal{M}$ as a binary mask that determines the quantization method assigned to each group.
To analyze its behavior, we randomly sample $50$ instances from the evaluation set of each task and record $\mathcal{M}$ for all layers.
Across tasks and models, $\mathcal{M}$ has an average sparsity of $99\%$.
The only exception is Llama 3.2-1B on Minerva Math, where the average sparsity drops to $90\%$.
Overall, these observations suggest that hybrid quantization overwhelmingly favors symmetric quantization across quantization groups.
However, as reported in Table~\ref{tab:accuracy}, this seemingly slight difference in the quantization mode results in an average $1.3$ score improvement for InnerQ\textsubscript{Hybrid} over InnerQ\textsubscript{Small}.

We also ablate the sparsity of $\mathcal{M}$ and observe its impact on the latency of the fused dequantization kernel.
Comparing the value cache latencies in Table~\ref{tab:hybrid_ablation} with the reported latencies in Table~\ref{tab:latencies_dequant} shows that reducing the sparsity from $99\%$ to $90\%$ slightly increases the latency of the fused kernel.
Further reducing the sparsity increases this latency to levels higher than the latency of InnerQ\textsubscript{Base}.
However, even at the lowest sparsity, the latency of InnerQ\textsubscript{Hybrid} is still lower than both TurboQuant and KIVI.

\subsection{Quantization Mode}
\label{sec:ablation_q_mode}

\input{table_ablation_q_mode}

We study the effect of varying the KV cache quantization mode on model evaluation performance.
We change the quantization mode of the key and value cache matrices between symmetric and asymmetric forms and report the results in Table~\ref{tab:ablation_q_mode}.

While we observe an overall decreasing trend in the model score when using asymmetric quantization, the extent of the score drop depends on the model and the setup.
When both the key and value caches are quantized to 3 bits, the effect of the quantization mode is more limited.
For example, for Llama 3.1-8B, a noticeable reduction appears only when both $K$ and $V$ are quantized asymmetrically, while for Llama 3.2-3B the model score drops as soon as either $K$ or $V$ is changed to asymmetric mode.
In contrast, when the $V$ bit-width is reduced to 2 bits, switching to asymmetric quantization has a larger impact on the evaluation score across all models.

The last row in each bit-width group applies hybrid quantization to $V$ to address the above-mentioned sensitivity of the KV cache to the quantization mode.
Hybrid quantization adjusts the quantization mode per-group according to each group's underlying data distributions.
The results show that this approach largely recovers the score reduction associated with asymmetric quantization and, in some cases, exceeds the best non-hybrid configuration.
This effect is more pronounced when the value cache is quantized to 2 bits.
For instance, in Llama 2-7B, the configuration with hybrid quantization for the value cache is the only one that maintains a non-trivial score, whereas the remaining configurations yield near-zero results.

Taken together, the results in Table~\ref{tab:ablation_q_mode} indicate that the choice of quantization mode significantly affects evaluation performance, and that the adaptive hybrid quantization provides an effective mechanism for preserving performance under low-bit KV cache quantization.

\section{Conclusion}

In this work, we introduced InnerQ, a hardware-aware KV cache quantization scheme that achieves an average speedup of $1.3\times$ over existing KV cache quantization methods and $2.7\times$ over the non-quantized cache.
This improvement is enabled by applying group-wise quantization with groups formed along the inner dimension of the quantized matrices.
By aligning the quantization layout with the execution flow of vector-matrix multiplication, InnerQ reduces memory traffic and dequantization overhead.

To preserve evaluation performance, InnerQ combines this hardware-friendly layout with high-precision windows for both recent tokens and early attention sinks, together with per-channel normalization of the key cache.
We introduced three InnerQ variants that offer different trade-offs between evaluation performance and inference latency: InnerQ\textsubscript{Base}, InnerQ\textsubscript{Hybrid}, and InnerQ\textsubscript{Small}.
InnerQ\textsubscript{Base} prioritizes metric score, InnerQ\textsubscript{Small} emphasizes latency reduction, and InnerQ\textsubscript{Hybrid} provides an intermediate trade-off.

Our results show that all InnerQ variants improve inference speed over prior KV cache quantization methods.
In particular, InnerQ\textsubscript{Base} achieves evaluation performance comparable to strong existing methods such as TurboQuant~\cite{turboquant} and KIVI~\cite{kivi}, while reducing KV cache latency.
The results further demonstrate the effectiveness of our proposed hybrid quantization strategy used in InnerQ\textsubscript{Hybrid}, which maintains competitive evaluation scores despite the low bit-width value cache.

Overall, InnerQ demonstrates that KV cache quantization can benefit substantially from co-designing numerical representation with hardware memory-access patterns.
These results highlight the importance of hardware-aware quantization layouts for maximizing efficiency gains in long-context large language model inference.

Future work will systematically examine how our quantization scheme interacts with alternative KV cache eviction strategies.
Another direction is to design adaptive and low-overhead policies for selecting tokens that are retained in the high-precision windows.

\bibliographystyle{abbrvnat}
\bibliography{references}

\medskip

\end{document}

%% file: figure_inner_outer.tex
\begin{figure}[t]
    \centering
    \begin{subfigure}[b]{0.49\textwidth}
        \input{./figure_inner_outer_tikz_outer}
        \caption{Grouping over the outer dimension}
        \label{fig:inner_outer:outer}
    \end{subfigure}
    \hfill
    \begin{subfigure}[b]{0.49\textwidth}
        \input{./figure_inner_outer_tikz_inner}
        \caption{Grouping over the inner dimension}
        \label{fig:inner_outer:inner}
    \end{subfigure}
    \caption{
        Visualization of the vector-matrix multiplication between the floating-point vector $Q$ and the quantized matrix $\hat{K}_\mathrm{cache}$ in an illustrative example.
        Cells with the same color are in the same quantization group and share a scale factor and zero point.
        A similar visualization is true for the vector $P$ and the quantized matrix $\hat{V}_\mathrm{cache}$.
    }
    \label{fig:inner_outer}
\end{figure}

%% file: figure_inner_outer_tikz_outer.tex
\resizebox{\linewidth}{!}{%
    \begin{tikzpicture}[
        square/.style={
            minimum size=0.6cm,
            outer sep=0pt,
            anchor=west,
            inner sep=0pt
        },
        label text/.style={
            font=\Large,
            anchor=south
        },
        arrow style/.style={
            ->,
            draw=teal!70!blue, 
            line width=2.0pt,
            >=stealth
        }
    ]
    \definecolor{colB1}{HTML}{4F81BD} \definecolor{colB2}{HTML}{E49F7A}
    \definecolor{colB3}{HTML}{54B153} \definecolor{colB4}{HTML}{D672D5}
    \definecolor{colB5}{HTML}{56DFC6} \definecolor{colB6}{HTML}{5E7D32}
    \definecolor{colB7}{HTML}{448A45} \definecolor{colB8}{HTML}{1F1F9F}
    \definecolor{colB9}{HTML}{7E3F1F} \definecolor{colB10}{HTML}{FFFF00}
    \definecolor{colB11}{HTML}{A6A6A6} \definecolor{colB12}{HTML}{386D29}
    \def\colors{colB1, colB2, colB3, colB4, colB5, colB6, colB7, colB8, colB9, colB10, colB11, colB12}

    \node[label text] at (-0.6, -0.3) {$Q$};
    \node[label text] at (3.6, -1.0) {$\times$};
    \node[label text] at (3.6, 0.4) {$d$};
    \node[label text] at (-0.7, -2.4) {$\hat{K}_\mathrm{cache}$};
    \foreach \i in {1,...,12} {
        \node[square, draw=black, fill=white] (A\i) at ({(\i-1)*0.6}, 0) {};
    }

    \begin{scope}[yshift=-1.5cm]
        \foreach \col [count=\i] in \colors {
            \node[square, fill=\col, draw=black] (B1_\i) at ({(\i-1)*0.6}, 0) {};
        }
        \foreach \col [count=\i] in \colors {
            \node[square, fill=\col, draw=black] (B2_\i) at ({(\i-1)*0.6}, -0.6) {};
        }
        \foreach \col [count=\i] in \colors {
            \node[square, fill=\col, draw=none, minimum height=.5cm, anchor=north west] 
                (B3_\i) at (B2_\i.south west) {};
            \draw[black, thin] (B3_\i.south west) -- (B3_\i.north west) -- (B3_\i.north east) -- (B3_\i.south east);
        }
        \node[fit=(B2_1) (B3_12), inner sep=0pt] (FadeArea) {};
        \fill[white, path fading=north] (FadeArea.north west) rectangle (FadeArea.south east);
        \node[draw=red!80!black, line width=3pt, inner sep=0pt, fit=(B1_1) (B1_12)] (RedBox) {};
    \end{scope}

    \coordinate (ConnectorX) at ($(B1_12.east) + (0.0, 0)$);
    \node[right=.5cm of ConnectorX, align=left, anchor=west] (Text) {For each row:\\$d$ scale factors\\$d$ zero-points\\\textbf{No reuse}};
    \draw[arrow style] (ConnectorX) -- (Text);
    \end{tikzpicture}
}

%% file: figure_inner_outer_tikz_inner.tex
\resizebox{\linewidth}{!}{%
    \begin{tikzpicture}[
        square/.style={
            minimum size=0.6cm,
            outer sep=0pt,
            anchor=west,
            inner sep=0pt
        },
        label text/.style={
            font=\Large,
            anchor=south
        },
        arrow style/.style={
            ->,
            draw=teal!70!blue, 
            line width=2.0pt,
            >=stealth
        }
    ]
    \definecolor{r1Left}{HTML}{4F81BD} 
    \definecolor{r1Right}{HTML}{E49F7A}
    \definecolor{r2Left}{HTML}{54B153}
    \definecolor{r2Right}{HTML}{D672D5}
    \definecolor{r3Left}{HTML}{56DFC6}
    \definecolor{r3Right}{HTML}{5E7D32}

    \node[label text] at (-0.6, -0.3) {$Q$};
    \node[label text] at (3.6, -1.0) {$\times$};
    \node[label text] at (3.6, 0.4) {$d$};
    \node[label text] at (-0.7, -2.4) {$\hat{K}_\mathrm{cache}$};
    \foreach \i in {1,...,12} {
        \node[square, draw=black, fill=white] (A\i) at ({(\i-1)*0.6}, 0) {};
    }

    \begin{scope}[yshift=-1.5cm]
        \foreach \i in {1,...,12} {
            \ifnum\i<7 \colorlet{curcol}{r1Left} \else \colorlet{curcol}{r1Right} \fi
            \node[square, fill=curcol, draw=black] (B1_\i) at ({(\i-1)*0.6}, 0) {};
        }

        \foreach \i in {1,...,12} {
            \ifnum\i<7 \colorlet{curcol}{r2Left} \else \colorlet{curcol}{r2Right} \fi
            \node[square, fill=curcol, draw=black] (B2_\i) at ({(\i-1)*0.6}, -0.6) {};
        }

        \foreach \i in {1,...,12} {
            \ifnum\i<7 \colorlet{curcol}{r3Left} \else \colorlet{curcol}{r3Right} \fi
            
            \node[square, fill=curcol, draw=none, minimum height=.5cm, anchor=north west] 
                (B3_\i) at (B2_\i.south west) {};
            
            \draw[black, thin] (B3_\i.south west) -- (B3_\i.north west) -- (B3_\i.north east) -- (B3_\i.south east);
        }

        \node[fit=(B2_1) (B3_12), inner sep=0pt] (FadeArea) {};
        \fill[white, path fading=north] (FadeArea.north west) rectangle (FadeArea.south east);
        
        \node[draw=red!80!black, line width=3pt, inner sep=0pt, fit=(B1_1) (B1_12)] (RedBox) {};
    \end{scope}
    \coordinate (ConnectorX) at ($(B1_12.east) + (0.0, 0)$);
    \node[right=.5cm of ConnectorX, align=left, anchor=west] (Text) {For each row:\\[0.1cm]$\frac{d}{G}$ scale factors\\[0.1cm]$\frac{d}{G}$ zero-points\\[0.1cm]\textbf{High reuse}};
    \draw[arrow style] (ConnectorX) -- (Text);
    \end{tikzpicture}%
}

%% file: figure_overview_quantization.tex
\begin{figure}[t]
    \centering
    \input{figure_overview_quantization_tikz}
    \caption{
        Overall schematics of using our proposed KV cache quantization method in the attention computation process (Equation~\ref{eq:att:1}).
        The value cache has a similar behavior.
        A small high-precision window consisting of sink tokens ($K_\mathrm{sink}$) and recent tokens ($K_\mathrm{recent}$) is maintained along with the quantized kernels ($\hat{K}$).
        The portions of the output corresponding to the quantized kernel and the full-precision windows are computed separately and then merged.
    }
    \label{fig:quantization_overall}
\end{figure}

%% file: figure_overview_quantization_tikz.tex
\resizebox{.8\linewidth}{!}{%
\begin{tikzpicture}[
    >=stealth,
    box/.style={
        rectangle, 
        draw=modernBlue, 
        fill=modernBlue,
        text=white,
        font=\small,
        minimum height=1cm
    },
    emptybox/.style={
        rectangle, 
        draw=black, 
        fill=none,
        text=black,
        font=\tiny,
        minimum height=0.5cm
    }
]

    \node (gemv) [box, minimum width=1cm] {GEMV};
    \node (dequant) [box, minimum width=1.5cm, below=0.7cm of gemv.south west, anchor=north west] {dequant};
    \node (qgemv) [box, minimum width=1cm, right=0.3cm of dequant.east, anchor=west] {GEMV};
    \node (fused) [
        draw, 
        dashed, 
        black!60, 
        fit=(dequant) (qgemv), 
        inner sep=0.1cm,
    ] {};
    \node (kcache) [emptybox, minimum width=4cm, left=0.6cm of dequant.south west, anchor=south east] {$\hat{K}_\mathrm{cache}$};
    \node (recents) [emptybox, minimum width=3cm, left=0.6cm of gemv.north west, anchor=north east] {$K_\mathrm{recent}$};
    \node (sinks) [emptybox, minimum width=1cm, left=0cm of recents.west, anchor=east] {$K_\mathrm{sink}$};
    \node (fusedtext) [
        above=0pt of fused.north west, 
        anchor=south west,
        font=\tiny,
        text=black!100
    ] {Fused Kernel};
    \coordinate (merge-top) at (gemv.north);
    \coordinate (merge-bottom) at (dequant.south);
    \node (merge) [
        box,
        fit=(merge-top) (merge-bottom), 
        inner sep=0pt,
        minimum width=0.8cm,
        right=0.6cm of qgemv.south east, 
        anchor=south west
    ] {};
    \node [font=\small, rotate=270, text=white] at (merge.center) {Concatenate};
    \coordinate (x-target) at ([xshift=-1cm]fused.west);
    \node (query) [font=\tiny] at (x-target |- merge.center) {$Q$};
    \node (output) [font=\tiny, right=0.4cm of merge.east, anchor=west] {$S$};

    \coordinate (x-accoladetext) at ([xshift=0.5cm,yshift=-0.5cm]recents.south west);
    \node (accoladetext) [font=\tiny, align=center, text width=3cm] at (x-accoladetext) {Old tokens to be\\quantized and moved to $\hat{K}_\mathrm{cache}$};
    \draw [decorate, 
           decoration={brace, amplitude=4pt, mirror, raise=2pt}, 
           thick] 
           (recents.south west) -- ++(1,0);

    \draw [->]
        (merge.east)
        -- (output.west);
    \draw [->] 
        (query.east)
        -- ++(0.3,0)
        -- ++(0,0.6) coordinate (p1)
        -- (p1 -| gemv.west);
    \draw [->] 
        (query.east)
        -- ++(0.3,0)
        -- ++(0,-0.6) coordinate (p2)
        -- (p2 -| fused.west);
    \draw [->] 
        (kcache.east)
        -- (kcache.east -| fused.west);
    \draw [->] 
        (recents.east)
        -- (recents.east -| gemv.west);
    \draw [->] 
        (dequant.east)
        -- (qgemv.west);
    \draw [->] 
        (fused.east)
        -- (fused.east -| merge.west);
    \draw [->] 
        (gemv.east) 
        -- (gemv.east -| merge.west);
\end{tikzpicture}
}

%% file: figure_hybrid.tex
\begin{figure}[t]
    \centering
    \input{figure_hybrid_tikz}
    \caption{
        High-level view of the quantization process in the hybrid quantization mode for a quantization group $\mathcal{G}_{i,g}$.
    }
    \label{fig:hybrid}
\end{figure}

%% file: figure_hybrid_tikz.tex
\resizebox{\linewidth}{!}{%
\begin{tikzpicture}[
    >=stealth,
    box/.style={
        rectangle, 
        draw=modernBlue, 
        fill=modernBlue,
        text=white,
        font=\tiny,
        minimum height=0.5cm,
        minimum width=1.1cm,
    },
    textbox/.style={
        rectangle, 
        text=black,
        font=\tiny,
    },
    mux/.style={
        trapezium,
        draw=modernBlue,
        fill=modernBlue,
        text=white,
        trapezium left angle=135,
        trapezium right angle=135, 
        minimum height=0.2cm,
        minimum width=1.7cm,
        font=\tiny,
        align=center
    }
]
    \node (quant_sym) [box] {quant\textsubscript{sym}};
    \node (quant_asym) [box, below=0.3cm of quant_sym.south, anchor=north] {quant\textsubscript{asym}};
    \node (dequant_sym) [box, right=2.8cm of quant_sym.east, anchor=west, minimum width=1.3cm] {dequant\textsubscript{sym}};
    \node (dequant_asym) [box, right=2.8cm of quant_asym.east, anchor=west, minimum width=1.3cm] {dequant\textsubscript{asym}};
    \node (quant_sym_out) at ($(quant_sym)!0.5!(dequant_sym)$) [textbox] {($\hat{\mathcal{G}}^\textrm{sym}_{i,g}$,$S^\textrm{sym}_{i,g}$)};
    \node (quant_asym_out) at ($(quant_asym)!0.5!(dequant_asym)$) [textbox] {($\hat{\mathcal{G}}^\textrm{asym}_{i,g}$,$S^\textrm{asym}_{i,g}$,$Z_{i,g}$)};
    \node (dequant_sym_out) [textbox,  right=0.2cm of dequant_sym.east, anchor=west] {$\bar{\mathcal{G}}^\textrm{sym}_{i,g}$};
    \node (dequant_asym_out) [textbox, right=0.2cm of dequant_asym.east, anchor=west] {$\bar{\mathcal{G}}^\textrm{asym}_{i,g}$};
    \node (g) at ([xshift=-0.5cm]quant_sym.west |- {$(quant_sym)!0.5!(quant_asym)$}) [textbox] {$\mathcal{G}_{i,g}$};

    \node (step1) [
        draw, 
        dashed, 
        black!60, 
        fit=(g) (quant_sym) (dequant_asym) (dequant_asym_out), 
        inner sep=0.1cm,
    ] {};
    \node (step1text) [
        above=0pt of step1.north west, 
        anchor=south west,
        font=\tiny,
        text=black
    ] {Step 1: Calculate Quantization Error};
    
    \draw [->] 
        (quant_sym.east) 
        -- (quant_sym_out.west);
    \draw [->] 
        (quant_asym.east) 
        -- (quant_asym_out.west);
    \draw [->] 
        (quant_sym_out.east) 
        -- (dequant_sym.west);
    \draw [->] 
        (quant_asym_out.east) 
        -- (dequant_asym.west);
    \draw [->] 
        (dequant_sym.east) 
        -- (dequant_sym_out.west);
    \draw [->] 
        (dequant_asym.east) 
        -- (dequant_asym_out.west);
    \draw [->] 
        (g.north) 
        -- (g.north |- quant_sym.west)
        -- (quant_sym.west);
    \draw [->] 
        (g.south) 
        -- (g.north |- quant_asym.west)
        -- (quant_asym.west);

    \node (sym_err) [textbox, right=.4cm of dequant_sym_out.east, anchor=west] {$|\mathcal{G}_{i,g}-\bar{\mathcal{G}}^\textrm{sym}_{i,g}|$};
    \node (asym_err) [textbox, right=.4cm of dequant_asym_out.east, anchor=west] {$|\mathcal{G}_{i,g}-\bar{\mathcal{G}}^\textrm{asym}_{i,g}|$};
    \coordinate (comp-top) at (dequant_sym.north);
    \coordinate (comp-bottom) at (dequant_asym.south);
    \node (comp) [
        box,
        fit=(comp-top) (comp-bottom), 
        inner sep=0pt,
        minimum width=0.8cm,
        right=0.3cm of asym_err.south east, 
        anchor=south west
    ] {};
    \node [font=\tiny, rotate=270, text=white] at (comp.center) {Comparator};
    \node (m) [textbox, right=0.2cm of comp.east, anchor=west] {$\mathcal{M}^\textrm{asym}_{i,g}$};
    \node (step2) [
        draw, 
        dashed, 
        black!60, 
        fit=(sym_err) (asym_err) (comp) (m), 
        inner sep=0.1cm,
    ] {};
    \node (step2text) [
        above=0pt of step2.north west, 
        anchor=south west,
        font=\tiny,
        text=black
    ] {Step 2: Choose Quantization Mode};
    \draw [->] 
        (sym_err.east) 
        -- (sym_err.east -| comp.west);
    \draw [->] 
        (asym_err.east) 
        -- (asym_err.east -| comp.west);
    \draw [->] 
        (comp.east) 
        -- (m.west);

    \node (mux1_sel) [textbox, below=1.2cm of g.south east, anchor=north west] {$\mathcal{M}^\textrm{asym}_{i,g}$};
    \node (mux1) [mux, right=.3cm of mux1_sel.east, anchor=west] {0 \hspace{1cm} 1};
    \node (mux1_inp1) at ([xshift=-.6cm,yshift=.2cm]mux1.north) [textbox] {$\hat{\mathcal{G}}^\textrm{sym}_{i,g}$};
    \node (mux1_inp2) at ([xshift=.6cm,yshift=.2cm]mux1.north) [textbox] {$\hat{\mathcal{G}}^\textrm{sym}_{i,g}$};
    \node (mux1_out) at ([xshift=0cm,yshift=-.5cm]mux1.south) [textbox] {$\hat{\mathcal{G}}^\textrm{hybrid}_{i,g}$};
    
    \node (mux2_sel) [textbox, right=0.5cm of mux1.east, anchor=west] {$\mathcal{M}^\textrm{asym}_{i,g}$};
    \node (mux2) [mux, right=.3cm of mux2_sel.east, anchor=west] {0 \hspace{1cm} 1};
    \node (mux2_inp1) at ([xshift=-.6cm,yshift=.2cm]mux2.north) [textbox] {$S^\textrm{sym}_{i,g}$};
    \node (mux2_inp2) at ([xshift=.6cm,yshift=.2cm]mux2.north) [textbox] {$-S^\textrm{asym}_{i,g}$};
    \node (mux2_out) at ([xshift=0cm,yshift=-.5cm]mux2.south) [textbox] {$S^\textrm{hybrid}_{i,g}$};
    
    \node (mux3_sel) [textbox, right=0.5cm of mux2.east, anchor=west] {$\mathcal{M}^\textrm{asym}_{i,g}$};
    \node (mux3) [mux, right=.3cm of mux3_sel.east, anchor=west] {0 \hspace{1cm} 1};
    \node (mux3_inp1) at ([xshift=-.6cm,yshift=.2cm]mux3.north) [textbox] {$0$};
    \node (mux3_inp2) at ([xshift=.6cm,yshift=.2cm]mux3.north) [textbox] {$Z_{i,g}$};
    \node (mux3_out) at ([xshift=0cm,yshift=-.5cm]mux3.south) [textbox] {$Z^\textrm{hybrid}_{i,g}$};
    \coordinate (helper) at ([xshift=-0.1cm] step2.east |- mux3.east);
    
    \node (step3) [
        draw, 
        dashed, 
        black!60, 
        fit=(mux1_sel) (mux1_inp1) (mux1_out) (helper), 
        inner sep=0.1cm,
    ] {};
    \node (step3text) [
        left=0pt of step3.south west, 
        anchor=south west,
        font=\tiny,
        rotate=90,
        align=left,
        text=black
    ] {Step 3:\\Prepare Outputs};
    \draw [->] 
        (mux1_sel.east) 
        -- (mux1_sel.east -| mux1.west);
    \draw [->] 
        (mux2_sel.east) 
        -- (mux2_sel.east -| mux2.west);
    \draw [->] 
        (mux3_sel.east) 
        -- (mux3_sel.east -| mux3.west);
    \draw [->] 
        (mux1.south) 
        -- (mux1_out.north);
    \draw [->] 
        (mux2.south) 
        -- (mux2_out.north);
    \draw [->] 
        (mux3.south) 
        -- (mux3_out.north);

\end{tikzpicture}
}

%% file: table_results_main.tex
\begin{table}[t]
\centering
\scriptsize
\begin{tabular}{cc|ccccccc}
\begin{tabular}{@{}c@{}}Dataset\end{tabular}&Model&Baseline&KIVI&KIVI\textsubscript{Sink}&TurboQuant&InnerQ\textsubscript{Base}&InnerQ\textsubscript{Hybrid}&InnerQ\textsubscript{Small}\\
\specialrule{1pt}{2pt}{2pt}
\multirow{6}{*}{GSM8k}&
Llama 3.2-1B&
$6.22$&$2.81$&$4.02$&$$5.84$$&$5.53$&$5.31$&$4.63$
\\
&Llama 3.2-3B&
$26.46$&$19.33$&$23.35$&$25.47$&$27.14$&$23.12$&$24.03$
\\
&Llama 3.1-8B&
$49.96$&$42.84$&$47.84$&$48.90$&$48.22$&$47.31$&$45.94$
\\
&Llama 2-7B&
$13.04$&$13.12$&$11.90$&$12.59$&$13.42$&$11.30$&$0.46$
\\
&Mistral 7B v0.3&
$36.92$&$32.30$&$33.81$&$34.72$&$36.47$&$33.89$&$33.89$
\\
\cmidrule{2-9}
&Average&
$26.52$&$22.08$&$24.18$&$25.50$&$26.16$&$24.19$&$21.79$
\\
\specialrule{0.5pt}{2pt}{2pt}
\multirow{6}{*}{\begin{tabular}[c]{@{}c@{}}Minerva\\Math (500)\end{tabular}}&
Llama 3.2-1B&
$4.40$&$5.20$&$3.80$&$3.60$&$3.40$&$3.40$&$2.80$
\\
&Llama 3.2-3B&
$9.20$&$6.80$&$8.60$&$11.40$&$10.80$&$8.40$&$6.60$
\\
&Llama 3.1-8B&
$21.60$&$17.20$&$21.00$&$21.20$&$20.80$&$19.60$&$19.00$
\\
&Llama 2-7B&
$3.20$&$3.60$&$4.00$&$5.20$&$3.00$&$3.40$&$0.40$
\\
&Mistral 7B v0.3&
$11.60$&$10.40$&$10.40$&$12.80$&$12.60$&$11.40$&$9.80$
\\
\cmidrule{2-9}
&Average&
$10.00$&$8.64$&$9.56$&$10.84$&$10.12$&$9.24$&$7.72$
\\
\specialrule{0.5pt}{2pt}{2pt}
\multirow{5}{*}{\begin{tabular}[c]{@{}c@{}}HumanEval\end{tabular}}&
Llama 3.2-1B&
$33.54$&$21.34$&$33.54$&$32.72$&$34.15$&$27.44$&$28.66$
\\
&Llama 3.2-3B&
$52.44$&$52.44$&$53.05$&$56.71$&$52.44$&$53.66$&$53.05$
\\
&Llama 3.1-8B&
$67.07$&$64.02$&$64.02$&$63.42$&$66.46$&$64.63$&$64.63$
\\
&Mistral 7B v0.3&
$40.85$&$41.46$&$36.59$&$41.46$&$44.51$&$40.85$&$39.02$
\\
\cmidrule{2-9}
&Average&
$48.47$&$44.81$&$46.80$&$48.58$&$49.39$&$46.64$&$46.34$
\\
\specialrule{0.5pt}{2pt}{2pt}
\multirow{5}{*}{\begin{tabular}[c]{@{}c@{}}MBPP\end{tabular}}&
Llama 3.2-1B&
$37.00$&$26.20$&$34.00$&$34.20$&$37.80$&$35.00$&$34.20$
\\
&Llama 3.2-3B&
$51.60$&$49.40$&$49.40$&$49.60$&$50.00$&$49.40$&$49.20$
\\
&Llama 3.1-8B&
$60.00$&$57.20$&$58.40$&$59.80$&$59.40$&$59.80$&$59.00$
\\
&Mistral 7B v0.3&
$42.60$&$40.40$&$39.20$&$42.60$&$42.40$&$39.80$&$38.80$
\\
\cmidrule{2-9}
&Average&
$47.80$&$43.30$&$45.25$&$46.55$&$47.40$&$46.00$&$45.30$
\\
\specialrule{1pt}{2pt}{2pt}
\end{tabular}
\caption{
Evaluation metric score of models on different language generation tasks with quantized KV cache, compared to the non-quantized \texttt{FP16} baseline.
We report \texttt{flexible\_extract} for GSM8k, \texttt{math\_verify} for Minerva Math, and \texttt{pass\_at\_1} for Humaneval and MBPP tasks.
We use instructtion-tuned models for the Humaneval and MBPP tasks.
}
\label{tab:accuracy}
\end{table}

%% file: table_results_longbench.tex
\begin{table}[t]
\centering
\footnotesize
\begin{tabular}{cc|cccccc}
\begin{tabular}{@{}c@{}}Dataset\end{tabular}&Model&Baseline&KIVI&KIVI\textsubscript{Sink}&InnerQ\textsubscript{Base}&InnerQ\textsubscript{Hybrid}&InnerQ\textsubscript{Small}\\
\specialrule{1pt}{2pt}{2pt}
\multirow{5}{*}{qasper}&
Llama 3.2-1B&
$20.29$&$18.00$&$17.23$&$18.37$&$20.43$&$19.84$
\\
&Llama 3.2-3B&
$40.63$&$37.09$&$38.49$&$40.32$&$39.59$&$38.84$
\\
&Llama 3.1-8B&
$44.58$&$44.08$&$43.61$&$44.59$&$42.40$&$42.36$
\\
&Llama 2-7B&
$20.99$&$21.11$&$20.30$&$21.84$&$22.18$&$13.12$
\\
&Llama 2-13B&
$16.59$&$15.09$&$15.48$&$17.85$&$16.56$&$17.50$
\\
\specialrule{0.5pt}{2pt}{2pt}
\multirow{5}{*}{gov-report}&
Llama 3.2-1B&
$29.27$&$28.57$&$28.87$&$28.89$&$26.72$&$26.70$
\\
&Llama 3.2-3B&
$33.86$&$32.11$&$33.76$&$33.38$&$32.73$&$32.25$
\\
&Llama 3.1-8B&
$34.61$&$34.47$&$34.33$&$34.33$&$33.62$&$33.80$
\\
&Llama 2-7B&
$26.88$&$26.68$&$26.64$&$26.12$&$25.18$&$17.15$
\\
&Llama 2-13B&
$27.82$&$27.01$&$27.10$&$27.29$&$26.38$&$25.69$
\\
\specialrule{0.5pt}{2pt}{2pt}
\multirow{5}{*}{multinews}&
Llama 3.2-1B&
$26.08$&$25.89$&$26.02$&$26.07$&$25.06$&$24.70$
\\
&Llama 3.2-3B&
$25.93$&$25.74$&$26.07$&$25.70$&$25.43$&$25.35$
\\
&Llama 3.1-8B&
$26.95$&$26.39$&$26.76$&$26.73$&$26.20$&$26.09$
\\
&Llama 2-7B&
$26.17$&$25.93$&$25.8$&$25.80$&$26.18$&$11.7$
\\
&Llama 2-13B&
$26.56$&$25.78$&$26.26$&$26.17$&$25.97$&$26.16$
\\
\specialrule{0.5pt}{2pt}{2pt}
\multirow{5}{*}{trec}&
Llama 3.2-1B&
$2.00$&$2.75$&$3.50$&$4.75$&$4.50$&$3.50$
\\
&Llama 3.2-3B&
$10.03$&$8.53$&$9.03$&$10.50$&$8.10$&$9.10$
\\
&Llama 3.1-8B&
$39.50$&$32.50$&$34.50$&$37.00$&$37.50$&$38.00$
\\
&Llama 2-7B&
$64.00$&$64.50$&$64.50$&$64.50$&$64.50$&$62.00$
\\
&Llama 2-13B&
$68.50$&$68.00$&$68.00$&$67.50$&$67.50$&$66.50$
\\
\specialrule{0.5pt}{2pt}{2pt}
\multirow{5}{*}{triviaqa}&
Llama 3.2-1B&
$64.03$&$63.92$&$64.75$&$62.92$&$59.06$&$59.20$
\\
&Llama 3.2-3B&
$69.39$&$67.29$&$67.53$&$67.48$&$65.37$&$65.41$
\\
&Llama 3.1-8B&
$91.64$&$90.53$&$91.33$&$91.51$&$91.70$&$91.70$
\\
&Llama 2-7B&
$83.51$&$82.77$&$82.27$&$83.84$&$83.32$&$39.51$
\\
&Llama 2-13B&
$87.85$&$87.97$&$87.83$&$87.47$&$88.27$&$84.55$
\\
\specialrule{0.5pt}{2pt}{2pt}
\multirow{5}{*}{samsum}&
Llama 3.2-1B&
$6.04$&$6.83$&$6.11$&$6.57$&$6.18$&$6.25$
\\
&Llama 3.2-3B&
$6.70$&$7.16$&$7.48$&$7.00$&$7.06$&$7.13$
\\
&Llama 3.1-8B&
$43.57$&$42.33$&$43.09$&$43.38$&$42.82$&$43.12$
\\
&Llama 2-7B&
$41.36$&$40.37$&$40.44$&$40.62$&$40.16$&$14.77$
\\
&Llama 2-13B&
$42.55$&$42.12$&$42.57$&$42.16$&$41.77$&$40.54$
\\
\specialrule{0.5pt}{2pt}{2pt}
\multirow{5}{*}{lcc}&
Llama 3.2-1B&
$19.29$&$19.17$&$18.70$&$18.29$&$18.15$&$18.53$
\\
&Llama 3.2-3B&
$31.65$&$31.27$&$31.01$&$30.20$&$30.98$&$31.63$
\\
&Llama 3.1-8B&
$43.02$&$42.85$&$42.66$&$42.97$&$42.76$&$42.84$
\\
&Llama 2-7B&
$58.29$&$57.79$&$58.12$&$58.41$&$57.80$&$40.41$
\\
&Llama 2-13B&
$48.22$&$47.31$&$48.51$&$48.04$&$49.50$&$46.45$
\\
\specialrule{0.5pt}{2pt}{2pt}
\multirow{5}{*}{repobench-p}&
Llama 3.2-1B&
$16.61$&$17.33$&$15.41$&$14.54$&$14.43$&$15.84$
\\
&Llama 3.2-3B&
$29.56$&$29.36$&$29.44$&$29.19$&$29.02$&$29.34$
\\
&Llama 3.1-8B&
$38.26$&$38.21$&$38.37$&$38.23$&$38.08$&$37.93$
\\
&Llama 2-7B&
$52.05$&$51.57$&$52.04$&$52.03$&$50.43$&$28.71$
\\
&Llama 2-13B&
$49.78$&$48.59$&$49.03$&$49.71$&$49.05$&$46.14$
\\
\specialrule{1pt}{2pt}{2pt}
\end{tabular}
\caption{
Evaluation metric score of models on select tasks from the longbench suite with quantized KV cache, compared to the non-quantized \texttt{FP16} baseline.
We report \texttt{F1} for qasper and triviaqa, \texttt{ROUGE-L} for gov-report, multinews, and samsum, accuracy for trec, and \texttt{pass\_at\_1} for repobench-p.
We use instruction-tuned models for Llama 3 models and chat version of Llama 2 models.
}
\label{tab:score_longbench}
\end{table}

%% file: table_bits.tex
\begin{table}[t]
\centering
\small
\begin{tabular}{cc|ccccc}
&&KIVI&TurboQuant&InnerQ\textsubscript{Base}&InnerQ\textsubscript{Hybrid}&InnerQ\textsubscript{Small}\\
\specialrule{1pt}{2pt}{2pt}
\multirow{4}{*}{\begin{tabular}[c]{@{}c@{}}Key\\Cache\end{tabular}}
&\begin{tabular}[c]{@{}c@{}}\vspace{1pt}Integer Bit-width\vspace{1pt}\end{tabular}&
$2$&$4$&$3$&$3$&$3$
\\
&\begin{tabular}[c]{@{}c@{}}\vspace{1pt}Scale Factor Overhead\vspace{1pt}\end{tabular}&
$0.5$&$-$&$0.5$&$0.5$&$0.5$
\\
&\begin{tabular}[c]{@{}c@{}}\vspace{1pt}Zero-point Overhead\vspace{1pt}\end{tabular}&
$0.5$&$-$&$-$&$-$&$-$
\\
&\begin{tabular}[c]{@{}c@{}}\vspace{1pt}Channel Norm Overhead\vspace{1pt}\end{tabular}&
$-$&$0.25$&$-$&$-$&$-$
\\
\specialrule{0.5pt}{2pt}{2pt}
\multirow{4}{*}{\begin{tabular}[c]{@{}c@{}}Value\\Cache\end{tabular}}
&\begin{tabular}[c]{@{}c@{}}\vspace{1pt}Integer Bit-width\vspace{1pt}\end{tabular}&
$2$&$3$&$3$&$2$&$2$
\\
&\begin{tabular}[c]{@{}c@{}}\vspace{1pt}Scale Factor Overhead\vspace{1pt}\end{tabular}&
$0.5$&$-$&$0.5$&$0.5$&$0.5$
\\
&\begin{tabular}[c]{@{}c@{}}\vspace{1pt}Zero-point Overhead\vspace{1pt}\end{tabular}&
$0.5$&$-$&$-$&$0.5$&$-$
\\
&\begin{tabular}[c]{@{}c@{}}\vspace{1pt}Channel Norm Overhead\vspace{1pt}\end{tabular}&
$-$&$0.25$&$-$&$-$&$-$
\\
\specialrule{0.5pt}{2pt}{2pt}
\multicolumn{2}{c|}{Per-number Effective Bit-width}&
$3$&$3.75$&$3.5$&$3.25$&$3$
\\
\specialrule{1pt}{2pt}{2pt}
\end{tabular}
\caption{
The per-number effective bit-width of different KV cache quantization methods.
Bits for scale factors and zero-points are averaged to represent the overhead on each quantized number.
We assume that scale factors and zero-points are represented using \texttt{FP16} while channel norms are represented using \texttt{FP32}.
The group size is $32$ and the head dimension is $128$.
}
\label{tab:bits}
\end{table}

%% file: table_latency_dequant.tex
\begin{table}[t]
\centering
\small
\begin{tabular}{cc|ccccccc}
&&\multicolumn{7}{c}{Sequence Length}\\
\cmidrule(l){3-9}
&Method&
512&1024&2048&4096&8192&16384&32768
\\
\specialrule{1pt}{2pt}{2pt}
\multirow{6}{*}{\begin{tabular}[c]{@{}c@{}}Key Cache\\(Equation~\eqref{eq:att:1})\end{tabular}}
&Baseline (\texttt{FP16})&
$76$&$147$&$291$&$576$&$1148$&$2291$&$4593$
\\
&KIVI&
$39$&$72$&$138$&$270$&$535$&$1063$&$2120$
\\
&TurboQuant&
$34$&$62$&$118$&$230$&$453$&$901$&$1796$
\\
&InnerQ\textsubscript{Base}&
$30$&$53$&$99$&$192$&$378$&$749$&$1492$
\\
&InnerQ\textsubscript{Hybrid}&
$30$&$53$&$99$&$192$&$378$&$749$&$1492$
\\
&InnerQ\textsubscript{Small}&
$30$&$53$&$99$&$192$&$378$&$749$&$1492$
\\
\specialrule{0.5pt}{2pt}{2pt}
\multirow{6}{*}{\begin{tabular}[c]{@{}c@{}}Value Cache\\(Equation~\eqref{eq:att:3})\end{tabular}}
&Baseline (\texttt{FP16})&
$76$&$148$&$291$&$597$&$1172$&$2347$&$4922$
\\
&KIVI&
$40$&$73$&$139$&$273$&$538$&$1079$&$2210$
\\
&TurboQuant&
$40$&$78$&$149$&$286$&$563$&$1126$&$2250$
\\
&InnerQ\textsubscript{Base}&
$34$&$65$&$120$&$228$&$443$&$883$&$1784$
\\
&InnerQ\textsubscript{Hybrid}&
$33$&$59$&$110$&$214$&$423$&$842$&$1688$
\\
&InnerQ\textsubscript{Small}&
$32$&$57$&$109$&$211$&$416$&$826$&$1644$
\\
\specialrule{0.5pt}{2pt}{2pt}
\multirow{6}{*}{Total}
&Baseline (\texttt{FP16})&
$153$&$295$&$582$&$1174$&$2320$&$4638$&$9516$
\\
&KIVI&
$79$&$146$&$278$&$543$&$1074$&$2142$&$4331$
\\
&TurboQuant&
$75$&$140$&$267$&$516$&$1017$&$2027$&$4046$
\\
&InnerQ\textsubscript{Base}&
$64$&$118$&$220$&$420$&$822$&$1633$&$3276$
\\
&InnerQ\textsubscript{Hybrid}&
$63$&$112$&$210$&$406$&$801$&$1591$&$3180$
\\
&InnerQ\textsubscript{Small}&
$62$&$110$&$208$&$403$&$795$&$1575$&$3136$
\\
\specialrule{1pt}{2pt}{2pt}

\end{tabular}
\caption{
Latency breakdown ($\textrm{\textmu}$s) of the fused dequantize-GEMV kernels for Llama 3.1-8B using different KV cache quantization methods along with the \texttt{FP16} GEMV as the baseline.
}
\label{tab:latencies_dequant}
\end{table}

%% file: graph_speedup_figure.tex
\begin{figure}[t]
    \centering
    \begin{subfigure}[b]{\textwidth}
        \centering
        \input{graph_speedup_legend}
    \end{subfigure}
    \begin{subfigure}[c]{0.04\textwidth}
        \input{graph_speedup_ylabel}
    \end{subfigure}%
    \begin{subfigure}[c]{0.32\textwidth}
        \input{graph_speedup_vs_fp}
    \end{subfigure}%
    \begin{subfigure}[c]{0.32\textwidth}
        \input{graph_speedup_vs_kivi}
    \end{subfigure}%
    \begin{subfigure}[c]{0.32\textwidth}
        \input{graph_speedup_vs_tq}
    \end{subfigure}
    \caption{
        Total speedup in the operations in Llama 3.1-8B associated with KV cache when using InnerQ variants for KV cache quantization, relative to (left) non-quantized \texttt{FP16}, (middle) KIVI~\cite{kivi}, and (right) TurboQuant~\cite{turboquant}.
    }
    \label{fig:speedups}
\end{figure}

%% file: graph_speedup_legend.tex
\begin{tikzpicture}[
    >=stealth,
    emptybox/.style={
        rectangle, 
        font=\small,
        minimum height=1cm
    },
    box/.style={
        rectangle, 
        draw=modernBlue, 
        fill=modernBlue,
        text=white,
        font=\small,
        minimum height=1cm
    },
    textbox/.style={
        rectangle, 
        text=black,
        font=\small,
    },
    legendtext/.style={
        rectangle, 
        text=black,
        font=\scriptsize,
    },
]
    \node (innerq_base) [legendtext] {InnerQ\textsubscript{Base}};
    \draw [modernBlue, thick] ([xshift=-0.6cm]innerq_base.west) -- ++(0.6,0);
    \node[modernBlue,scale=0.8] at ([xshift=-0.3cm]innerq_base.west) {\pgfuseplotmark{*}};
    
    \node (innerq_hybrid) [legendtext, right=0.9 of innerq_base.east] {InnerQ\textsubscript{Hybrid}};
    \draw [modernPurple, thick] ([xshift=-0.6cm]innerq_hybrid.west) -- ++(0.6,0);
    \node[modernPurple,scale=0.8] at ([xshift=-0.3cm]innerq_hybrid.west) {\pgfuseplotmark{+}};
    
    \node (innerq_fast) [legendtext, right=0.9 of innerq_hybrid.east] {InnerQ\textsubscript{Small}};
    \draw [modernGreen, thick] ([xshift=-0.6cm]innerq_fast.west) -- ++(0.6,0);
    \node[modernGreen,scale=0.8] at ([xshift=-0.3cm]innerq_fast.west) {\pgfuseplotmark{o}};

    \coordinate (helper) at ([xshift=-0.7cm] innerq_base.north west);
    \coordinate (helper2) at ([xshift=0.1cm] innerq_fast.south east);
    
    \node (step3) [
        draw, 
        fit=(helper) (helper2), 
        inner sep=0.05cm,
    ] {};
\end{tikzpicture}

%% file: graph_speedup_ylabel.tex
\begin{tikzpicture}
    \def\myWidth{\textwidth}
    \def\myHeight{2.1cm}
    
    \def\xPos{0.0} 
    \def\yPos{0.7}
    \def\myRotation{90}
    
    \useasboundingbox (0,0) rectangle (\myWidth, \myHeight);
    
    \node[rotate=\myRotation, font=\footnotesize] 
        at (\xPos*\myWidth, \yPos*\myHeight) {Speedup ($\times$)};
        
\end{tikzpicture}

%% file: graph_speedup_vs_fp.tex
\begin{tikzpicture}
    \begin{axis}[
        width=0.7\textwidth, height=3.5cm,
        scale only axis,
        font=\footnotesize,
        grid=major,
        grid style={dashed, gray!30},
        xmode=log,
        log basis x={2},
        xtick={512,1024,2048,4096,8192,16384,32768},
        xticklabels={512,1024,2048,4096,8192,16384,32768},
        xticklabel style={rotate=90, anchor=east},
        ytick distance={0.5},
        ylabel style={align=center},
        y tick label style={
            /pgf/number format/.cd,
            fixed,
            fixed zerofill,
            precision=1,
        },
        xlabel={Sequence Length},
        title={Relative to \texttt{FP16}},
        title style={align=center},
        legend pos=south east,
        legend cell align={left},
        legend style={font=\scriptsize},
        ymin=1,
        ymax=3.5,
    ]
    \addplot[
        color=modernBlue,
        mark=*,
        semithick,
        mark size=1pt
    ]
    coordinates {
    (512,2.382)(1024,2.500)(2048,2.645)(4096,2.790)(8192,2.823)(16384,2.840)(32768,2.904)
    };
    
    \addplot[
        color=modernPurple,
        mark=+,
        semithick,
        mark size=1pt
    ]
    coordinates {
    (512,2.412)(1024,2.632)(2048,2.770)(4096,2.885)(8192,2.894)(16384,2.915)(32768,2.992)
    };

    \addplot[
        color=modernGreen,
        mark=*,
        semithick,
        mark size=1pt
    ]
    coordinates {
    (512,2.464)(1024,2.671)(2048,2.792)(4096,2.906)(8192,2.918)(16384,2.944)(32768,3.034)
    };
    \end{axis}
\end{tikzpicture}

%% file: graph_speedup_vs_kivi.tex
\begin{tikzpicture}
    \begin{axis}[
        width=0.7\textwidth, height=3.5cm,
        scale only axis,
        font=\footnotesize,
        grid=major,
        grid style={dashed, gray!30},
        xmode=log,
        log basis x={2},
        xtick={512,1024,2048,4096,8192,16384,32768},
        xticklabels={512,1024,2048,4096,8192,16384,32768},
        xticklabel style={rotate=90, anchor=east},
        ytick distance={0.1},
        ylabel style={align=center},
        y tick label style={
            /pgf/number format/.cd,
            fixed,
            fixed zerofill,
            precision=1,
        },
        xlabel={Sequence Length},
        title={Relative to KIVI},
        title style={align=center},
        legend pos=south east,
        legend cell align={left},
        legend style={font=\scriptsize},
        ymin=1,
        ymax=1.4,
    ]
    \addplot[
        color=modernBlue,
        mark=*,
        semithick,
        mark size=1pt
    ]
    coordinates {
    (512,1.233)(1024,1.234)(2048,1.265)(4096,1.292)(8192,1.307)(16384,1.312)(32768,1.322)
    };
    
    \addplot[
        color=modernPurple,
        mark=+,
        semithick,
        mark size=1pt
    ]
    coordinates {
    (512,1.248)(1024,1.299)(2048,1.324)(4096,1.336)(8192,1.340)(16384,1.346)(32768,1.362)
    };

    \addplot[
        color=modernGreen,
        mark=*,
        semithick,
        mark size=1pt
    ]
    coordinates {
    (512,1.275)(1024,1.319)(2048,1.335)(4096,1.346)(8192,1.351)(16384,1.360)(32768,1.381)
    };
    \end{axis}
\end{tikzpicture}

%% file: graph_speedup_vs_tq.tex
\begin{tikzpicture}
    \begin{axis}[
        width=0.7\textwidth, height=3.5cm,
        scale only axis,
        font=\footnotesize,
        grid=major,
        grid style={dashed, gray!30},
        xmode=log,
        log basis x={2},
        xtick={512,1024,2048,4096,8192,16384,32768},
        xticklabels={512,1024,2048,4096,8192,16384,32768},
        xticklabel style={rotate=90, anchor=east},
        ytick distance={0.1},
        ylabel style={align=center},
        y tick label style={
            /pgf/number format/.cd,
            fixed,
            fixed zerofill,
            precision=1,
        },
        xlabel={Sequence Length},
        title={Relative to TurboQuant},
        title style={align=center},
        legend pos=south east,
        legend cell align={left},
        legend style={font=\scriptsize},
        ymin=1,
        ymax=1.4,
    ]
    \addplot[
        color=modernBlue,
        mark=*,
        semithick,
        mark size=1pt
    ]
    coordinates {
    (512,1.167)(1024,1.187)(2048,1.214)(4096,1.228)(8192,1.238)(16384,1.241)(32768,1.235)
    };
    
    \addplot[
        color=modernPurple,
        mark=+,
        semithick,
        mark size=1pt
    ]
    coordinates {
    (512,1.181)(1024,1.250)(2048,1.271)(4096,1.270)(8192,1.269)(16384,1.274)(32768,1.272)
    };

    \addplot[
        color=modernGreen,
        mark=*,
        semithick,
        mark size=1pt
    ]
    coordinates {
    (512,1.207)(1024,1.269)(2048,1.281)(4096,1.280)(8192,1.280)(16384,1.287)(32768,1.290)
    };
    \end{axis}
\end{tikzpicture}

%% file: table_latency_quant.tex
\begin{table}[t]
\centering
\small
\begin{tabular}{c|ccccc}
&KIVI&TurboQuant&InnerQ\textsubscript{Base}&InnerQ\textsubscript{Hybrid}&InnerQ\textsubscript{Small}\\
\specialrule{1pt}{2pt}{2pt}
Key Cache
&$1.0$&$15.9$&$17.3$&$17.3$&$17.3$
\\
\specialrule{0.5pt}{2pt}{2pt}
Value Cache
&$21.1$&$16.1$&$1.0$&$1.4$&$0.9$
\\
\specialrule{0.5pt}{2pt}{2pt}
Total
&$22.1$&$31.9$&$18.3$&$18.7$&$18.2$
\\
\specialrule{1pt}{2pt}{2pt}
\end{tabular}
\caption{
Latency breakdown ($\textrm{\textmu}$s) of the quantization operations during the decode phase for Llama 3.1-8B using different quantization methods.
}
\label{tab:latencies_quant}
\end{table}

%% file: graph_ablation_figure.tex
\begin{figure}[t]
    \centering
    \begin{subfigure}[b]{\textwidth}
        \centering
        \input{graph_ablation_legend}
    \end{subfigure}
    \begin{subfigure}[c]{\textwidth}
        \begin{subfigure}[c]{0.04\textwidth}
            \input{graph_ablation_ylabel}
        \end{subfigure}%
        \begin{subfigure}[c]{0.24\textwidth}
            \input{graph_ablation_llama_gsm8k}
        \end{subfigure}%
        \begin{subfigure}[c]{0.24\textwidth}
            \input{graph_ablation_llama_minerva}
        \end{subfigure}%
        \begin{subfigure}[c]{0.24\textwidth}
            \input{graph_ablation_llama_humaneval}
        \end{subfigure}%
        \begin{subfigure}[c]{0.24\textwidth}
            \input{graph_ablation_llama_mbpp}
        \end{subfigure}%
        \caption{
            Llama 3.1-8B
        } 
    \end{subfigure}
    \begin{subfigure}[c]{\textwidth}
        \begin{subfigure}[c]{0.04\textwidth}
            \input{graph_ablation_ylabel}
        \end{subfigure}%
        \begin{subfigure}[c]{0.24\textwidth}
            \input{graph_ablation_mistral_gsm8k}
        \end{subfigure}%
        \begin{subfigure}[c]{0.24\textwidth}
            \input{graph_ablation_mistral_minerva}
        \end{subfigure}%
        \begin{subfigure}[c]{0.24\textwidth}
            \input{graph_ablation_mistral_humaneval}
        \end{subfigure}%
        \begin{subfigure}[c]{0.24\textwidth}
            \input{graph_ablation_mistral_mbpp}
        \end{subfigure}%
        \caption{
            Mistral 7B v0.3
        } %
    \end{subfigure}%
    \caption{
        Effect of changing $w_\mathrm{sink}$ on the evaluation metric performance of (a) Llama 3.1-8B and (b) Mistral 7B v0.3.
        At each step $w_\mathrm{recent}=128 - w_\mathrm{sink}$.
        We report \texttt{flexible\_extract} for GSM8k, \texttt{math\_verify} for Minerva Math, and \texttt{pass\_at\_1} for Humaneval and MBPP tasks.
        We use instruct tuned models for the Humaneval and MBPP tasks.
    }
    \label{fig:ablation:sink}
\end{figure}

%% file: graph_ablation_legend.tex
\begin{tikzpicture}[
    >=stealth,
    emptybox/.style={
        rectangle, 
        font=\small,
        minimum height=1cm
    },
    box/.style={
        rectangle, 
        draw=modernBlue, 
        fill=modernBlue,
        text=white,
        font=\small,
        minimum height=1cm
    },
    textbox/.style={
        rectangle, 
        text=black,
        font=\small,
    },
    legendtext/.style={
        rectangle, 
        text=black,
        font=\scriptsize,
    },
]
    \node (kivi) [legendtext] {KIVI};
    \draw [modernRed, thick] ([xshift=-0.6cm]kivi.west) -- ++(0.6,0);
    \node[modernRed,scale=0.8] at ([xshift=-0.3cm]kivi.west) {\pgfuseplotmark{square*}};
    
    \node (innerq_base) [legendtext, right=0.9 of kivi.east] {InnerQ\textsubscript{Base}};
    \draw [modernBlue, thick] ([xshift=-0.6cm]innerq_base.west) -- ++(0.6,0);
    \node[modernBlue,scale=0.8] at ([xshift=-0.3cm]innerq_base.west) {\pgfuseplotmark{*}};
    
    \node (innerq_hybrid) [legendtext, right=0.9 of innerq_base.east] {InnerQ\textsubscript{Hybrid}};
    \draw [modernPurple, thick] ([xshift=-0.6cm]innerq_hybrid.west) -- ++(0.6,0);
    \node[modernPurple,scale=0.8] at ([xshift=-0.3cm]innerq_hybrid.west) {\pgfuseplotmark{+}};
    
    \node (innerq_fast) [legendtext, right=0.9 of innerq_hybrid.east] {InnerQ\textsubscript{Small}};
    \draw [modernGreen, thick] ([xshift=-0.6cm]innerq_fast.west) -- ++(0.6,0);
    \node[modernGreen,scale=0.8] at ([xshift=-0.3cm]innerq_fast.west) {\pgfuseplotmark{o}};

    \coordinate (helper) at ([xshift=-0.7cm] kivi.north west);
    \coordinate (helper2) at ([xshift=0.1cm] innerq_fast.south east);
    
    \node (step3) [
        draw, 
        fit=(helper) (helper2), 
        inner sep=0.05cm,
    ] {};
\end{tikzpicture}

%% file: graph_ablation_ylabel.tex
\begin{tikzpicture}
    \def\myWidth{\textwidth}
    \def\myHeight{0.1\textwidth}
    
    \def\xPos{0.0} 
    \def\yPos{0.0}
    \def\myRotation{90}
    
    \useasboundingbox (0,0) rectangle (\myWidth, \myHeight);
    
    \node[rotate=\myRotation, font=\footnotesize] 
        at (\xPos*\myWidth, \yPos*\myHeight) {Evaluation Score};
\end{tikzpicture}

%% file: graph_ablation_llama_gsm8k.tex
\begin{tikzpicture}
    \begin{axis}[
        width=0.7\textwidth, height=0.7\textwidth,
        scale only axis,
        font=\scriptsize,
        grid=major,
        grid style={dashed, gray!30},
        xtick={0,32,64,96},
        xtick style={draw=none},
        ytick style={draw=none},
        ylabel={},
        ylabel style={align=center},
        xlabel={$w_\mathrm{sink}$},
        title={GSM8k},
        title style={align=center},
        legend pos=south east,
        legend cell align={left},
        legend style={font=\scriptsize},
        outer sep=0pt,
    ]
    \addplot[
        color=modernRed,
        mark=square*,
        semithick,
        mark size=1pt
    ]
    coordinates {
        (0,42.835)(32,47.839)(64,47.46)(96,46.626)
    };
    \addplot[
        color=modernBlue,
        mark=*,
        semithick,
        mark size=1pt
    ]
    coordinates {
    (0,49.052)(32,48.218)(64,49.052)(96,47.309)
    };
    \addplot[
        color=modernPurple,
        mark=+,
        semithick,
        mark size=1pt
    ]
    coordinates {
    (0,39.727)(32,45.944)(64,44.276)(96,43.29)
    };
    \addplot[
        color=modernGreen,
        mark=o,
        semithick,
        mark size=1pt
    ]
    coordinates {
    (0,41.243)(32,47.309)(64,47.081)(96,43.139)
    };
    \end{axis}
\end{tikzpicture}

%% file: graph_ablation_llama_minerva.tex
\begin{tikzpicture}
    \begin{axis}[
        width=0.7\textwidth, height=0.7\textwidth,
        scale only axis,
        font=\scriptsize,
        grid=major,
        grid style={dashed, gray!30},
        xtick={0,32,64,96},
        xtick style={draw=none},
        ytick style={draw=none},
        ylabel={},
        ylabel style={align=center},
        xlabel={$w_\mathrm{sink}$},
        title={Minerva Math},
        title style={align=center},
        legend pos=south east,
        legend cell align={left},
        legend style={font=\scriptsize},
        outer sep=0pt,
    ]
    \addplot[
        color=modernRed,
        mark=square*,
        semithick,
        mark size=1pt
    ]
    coordinates {
    (0,17.2)(32,21)(64,20)(96,19.4)
    };
    \addplot[
        color=modernBlue,
        mark=*,
        semithick,
        mark size=1pt
    ]
    coordinates {
    (0,20.6)(32,20.8)(64,17.6)(96,20.2)
    };
    \addplot[
        color=modernPurple,
        mark=+,
        semithick,
        mark size=1pt
    ]
    coordinates {
    (0,11.4)(32,19)(64,18.2)(96,18.6)
    };
    \addplot[
        color=modernGreen,
        mark=o,
        semithick,
        mark size=1pt
    ]
    coordinates {
    (0,12.2)(32,19.6)(64,18.8)(96,19.4)
    };
    \end{axis}
\end{tikzpicture}

%% file: graph_ablation_llama_humaneval.tex
\begin{tikzpicture}
    \begin{axis}[
        width=0.7\textwidth, height=0.7\textwidth,
        scale only axis,
        font=\scriptsize,
        grid=major,
        grid style={dashed, gray!30},
        xtick={0,32,64,96},
        xtick style={draw=none},
        ytick style={draw=none},
        ylabel={},
        ylabel style={align=center},
        xlabel={$w_\mathrm{sink}$},
        title={Humaneval},
        title style={align=center},
        legend pos=south east,
        legend cell align={left},
        legend style={font=\scriptsize},
        outer sep=0pt,
    ]
    \addplot[
        color=modernRed,
        mark=square*,
        semithick,
        mark size=1pt
    ]
    coordinates {
    (0,64.024)(32,64.024)(64,59.146)(96,63.415)
    };
    \addplot[
        color=modernBlue,
        mark=*,
        semithick,
        mark size=1pt
    ]
    coordinates {
    (0,62.805)(32,66.463)(64,67.073)(96,65.244)
    };
    \addplot[
        color=modernPurple,
        mark=+,
        semithick,
        mark size=1pt
    ]
    coordinates {
    (0,50)(32,64.634)(64,64.024)(96,62.195)
    };
    \addplot[
        color=modernGreen,
        mark=o,
        semithick,
        mark size=1pt
    ]
    coordinates {
    (0,51.829)(32,64.634)(64,65.854)(96,65.244)
    };
    \end{axis}
\end{tikzpicture}

%% file: graph_ablation_llama_mbpp.tex
\begin{tikzpicture}
    \begin{axis}[
        width=0.7\textwidth, height=0.7\textwidth,
        scale only axis,
        font=\scriptsize,
        grid=major,
        grid style={dashed, gray!30},
        xtick={0,32,64,96},
        xtick style={draw=none},
        ytick style={draw=none},
        ylabel={},
        ylabel style={align=center},
        xlabel={$w_\mathrm{sink}$},
        title={MBPP},
        title style={align=center},
        legend pos=south east,
        legend cell align={left},
        legend style={font=\scriptsize},
        outer sep=0pt,
    ]
    \addplot[
        color=modernRed,
        mark=square*,
        semithick,
        mark size=1pt
    ]
    coordinates {
    (0,57.2)(32,58.4)(64,58.2)(96,58.2)
    };
    \addplot[
        color=modernBlue,
        mark=*,
        semithick,
        mark size=1pt
    ]
    coordinates {
    (0,58.4)(32,59.4)(64,58.8)(96,58.8)
    };
    \addplot[
        color=modernPurple,
        mark=+,
        semithick,
        mark size=1pt
    ]
    coordinates {
    (0,48.8)(32,59)(64,57.4)(96,56.8)
    };
    \addplot[
        color=modernGreen,
        mark=o,
        semithick,
        mark size=1pt
    ]
    coordinates {
    (0,49)(32,59.8)(64,57.2)(96,57.4)
    };
    \end{axis}
\end{tikzpicture}

%% file: graph_ablation_mistral_gsm8k.tex
\begin{tikzpicture}
    \begin{axis}[
        width=0.7\textwidth, height=0.7\textwidth,
        scale only axis,
        font=\scriptsize,
        grid=major,
        grid style={dashed, gray!30},
        xtick={0,32,64,96},
        xtick style={draw=none},
        ytick style={draw=none},
        ylabel={},
        ylabel style={align=center},
        xlabel={$w_\mathrm{sink}$},
        title={GSM8k},
        title style={align=center},
        legend pos=south east,
        legend cell align={left},
        legend style={font=\scriptsize},
        outer sep=0pt,
    ]
    \addplot[
        color=modernRed,
        mark=square*,
        semithick,
        mark size=1pt
    ]
    coordinates {
    (0,32.297)(32,33.813)(64,33.889)(96,33.51) 
    };
    \addplot[
        color=modernBlue,
        mark=*,
        semithick,
        mark size=1pt
    ]
    coordinates {
    (0,36.694)(32,36.467)(64,37.528)(96,35.709)
    };
    \addplot[
        color=modernPurple,
        mark=+,
        semithick,
        mark size=1pt
    ]
    coordinates {
    (0,21.304)(32,33.889)(64,33.51)(96,33.738)
    };
    \addplot[
        color=modernGreen,
        mark=o,
        semithick,
        mark size=1pt
    ]
    coordinates {
    (0,19.864)(32,33.889)(64,36.088)(96,33.434)
    };
    \end{axis}
\end{tikzpicture}

%% file: graph_ablation_mistral_minerva.tex
\begin{tikzpicture}
    \begin{axis}[
        width=0.7\textwidth, height=0.7\textwidth,
        scale only axis,
        font=\scriptsize,
        grid=major,
        grid style={dashed, gray!30},
        xtick={0,32,64,96},
        xtick style={draw=none},
        ytick style={draw=none},
        ylabel={},
        ylabel style={align=center},
        xlabel={$w_\mathrm{sink}$},
        title={Minerva Math},
        title style={align=center},
        legend pos=south east,
        legend cell align={left},
        legend style={font=\scriptsize},
        outer sep=0pt,
    ]
    \addplot[
        color=modernRed,
        mark=square*,
        semithick,
        mark size=1pt
    ]
    coordinates {
    (0,10.4)(32,10.4)(64,11.4)(96,12)
    };
    \addplot[
        color=modernBlue,
        mark=*,
        semithick,
        mark size=1pt
    ]
    coordinates {
    (0,10.2)(32,12.6)(64,10.4)(96,11.8)
    };
    \addplot[
        color=modernPurple,
        mark=+,
        semithick,
        mark size=1pt
    ]
    coordinates {
    (0,8.4)(32,9.8)(64,12.4)(96,11.2)
    };
    \addplot[
        color=modernGreen,
        mark=o,
        semithick,
        mark size=1pt
    ]
    coordinates {
    (0,6.4)(32,11.4)(64,12.4)(96,9.8)
    };
    \end{axis}
\end{tikzpicture}

%% file: graph_ablation_mistral_humaneval.tex
\begin{tikzpicture}
    \begin{axis}[
        width=0.7\textwidth, height=0.7\textwidth,
        scale only axis,
        font=\scriptsize,
        grid=major,
        grid style={dashed, gray!30},
        xtick={0,32,64,96},
        xtick style={draw=none},
        ytick style={draw=none},
        ylabel={},
        ylabel style={align=center},
        xlabel={$w_\mathrm{sink}$},
        title={Humaneval},
        title style={align=center},
        legend pos=south east,
        legend cell align={left},
        legend style={font=\scriptsize},
        outer sep=0pt,
    ]
    \addplot[
        color=modernRed,
        mark=square*,
        semithick,
        mark size=1pt
    ]
    coordinates {
    (0,41.463)(32,36.585)(64,37.805)(96,40.854)
    };
    \addplot[
        color=modernBlue,
        mark=*,
        semithick,
        mark size=1pt
    ]
    coordinates {
    (0,37.195)(32,44.512)(64,43.293)(96,42.073)
    };
    \addplot[
        color=modernPurple,
        mark=+,
        semithick,
        mark size=1pt
    ]
    coordinates {
    (0,31.707)(32,39.024)(64,36.585)(96,40.854)
    };
    \addplot[
        color=modernGreen,
        mark=o,
        semithick,
        mark size=1pt
    ]
    coordinates {
    (0,31.098)(32,40.854)(64,39.634)(96,38.415)
    };
    \end{axis}
\end{tikzpicture}

%% file: graph_ablation_mistral_mbpp.tex
\begin{tikzpicture}
    \begin{axis}[
        width=0.7\textwidth, height=0.7\textwidth,
        scale only axis,
        font=\scriptsize,
        grid=major,
        grid style={dashed, gray!30},
        xtick={0,32,64,96},
        xtick style={draw=none},
        ytick style={draw=none},
        ylabel={},
        ylabel style={align=center},
        xlabel={$w_\mathrm{sink}$},
        title={MBPP},
        title style={align=center},
        legend pos=south east,
        legend cell align={left},
        legend style={font=\scriptsize},
        outer sep=0pt,
    ]
    \addplot[
        color=modernRed,
        mark=square*,
        semithick,
        mark size=1pt
    ]
    coordinates {
    (0,40.4)(32,39.2)(64,39.8)(96,39.8)
    };
    \addplot[
        color=modernBlue,
        mark=*,
        semithick,
        mark size=1pt
    ]
    coordinates {
    (0,37)(32,42.4)(64,42.2)(96,42)
    };
    \addplot[
        color=modernPurple,
        mark=+,
        semithick,
        mark size=1pt
    ]
    coordinates {
    (0,31)(32,38.8)(64,39.8)(96,38.6)
    };
    \addplot[
        color=modernGreen,
        mark=o,
        semithick,
        mark size=1pt
    ]
    coordinates {
    (0,32)(32,39.8)(64,38.8)(96,39)
    };
    \end{axis}
\end{tikzpicture}

%% file: table_ablation_hybrid.tex
\begin{table}[t]
\centering
\small
\begin{tabular}{c|cccc}
\multirow{2}{*}{Sparsity of $\mathcal{M}$} & \multicolumn{4}{c}{Sequence Length}\\
\cmidrule{2-5}
&1024&4096&16384&32768\\
\specialrule{1pt}{2pt}{2pt}
$99\%$&
$59.0$&$214.4$&$841.9$&$1685.4$
\\
$90\%$&
$61.2$&$218.6$&$849.0$&$1701.5$
\\
$50\%$&
$65.3$&$231.2$&$900.1$&$1800.7$
\\
$1\%$&
$65.9$&$233.1$&$910.1$&$1814.9$
\\
\specialrule{1pt}{2pt}{2pt}
\end{tabular}
\caption{
Latency ($\textrm{\textmu}$s) of the fused hybrid dequantization and GEMV in Llama 3.1-8B for Equation~\eqref{eq:att:3} for different cache sizes and levels of sparsity of $\mathcal{M}$.
}
\label{tab:hybrid_ablation}
\end{table}

%% file: table_ablation_q_mode.tex
\begin{table}[t]
\centering
\small
\begin{tabular}{cc|cccc}
\multirow{2}{*}{Bit-width} & \multirow{2}{*}{\begin{tabular}[c]{@{}c@{}}Quantization\\Mode\end{tabular}} & \multicolumn{4}{c}{Model}
\\
\cmidrule{3-6}
&&Llama 3.2-3B&Llama 3.1-8B&Llama 2-7B&Mistral 7B v0.3
\\
\specialrule{1pt}{2pt}{2pt}
\multirow{5}{*}{$K$:3,$V$:3}&
$K$:Sym,$V$:Sym&
$27.14$&$48.22$&$13.42$&$36.47$
\\
&
$K$:Sym,$V$:Asym&
$25.78$&$48.67$&$12.66$&$35.86$
\\
&
$K$:Asym,$V$:Sym&
$25.17$&$48.60$&$13.27$&$36.39$
\\
&
$K$:Asym,$V$:Asym&
$23.88$&$47.23$&$12.13$&$32.37$
\\
\cmidrule(lr){2-6}
&
$K$:Sym,$V$:Hybrid&
$27.29$&$48.60$&$13.27$&$37.60$
\\
\specialrule{0.5pt}{2pt}{2pt}
\multirow{5}{*}{$K$:3,$V$:2}&
$K$:Sym,$V$:Sym&
$24.03$&$45.94$&$0.45$&$33.89$
\\
&
$K$:Sym,$V$:Asym&
$14.03$&$36.69$&$3.41$&$2.05$
\\
&
$K$:Asym,$V$:Sym&
$21.76$&$44.05$&$0.68$&$33.36$
\\
&
$K$:Asym,$V$:Asym&
$10.31$&$30.93$&$2.50$&$1.67$
\\
\cmidrule(lr){2-6}
&
$K$:Sym,$V$:Hybrid&
$23.12$&$47.31$&$11.30$&$33.89$
\\
\specialrule{1pt}{2pt}{2pt}
\end{tabular}
\caption{
\texttt{flexible\_extract} evaluation score on GSM8K for different KV-cache quantization modes across four language models.
In all configurations, both $K$ and $V$ are quantized along their inner dimensions.
}
\label{tab:ablation_q_mode}
\end{table}

%% file: neurips_2025.bbl
\begin{thebibliography}{23}
\providecommand{\natexlab}[1]{#1}
\providecommand{\url}[1]{\texttt{#1}}
\expandafter\ifx\csname urlstyle\endcsname\relax
  \providecommand{\doi}[1]{doi: #1}\else
  \providecommand{\doi}{doi: \begingroup \urlstyle{rm}\Url}\fi

\bibitem[Austin et~al.(2021)Austin, Odena, Nye, Bosma, Michalewski, Dohan,
  Jiang, Cai, Terry, Le, et~al.]{mbpp}
J.~Austin, A.~Odena, M.~Nye, M.~Bosma, H.~Michalewski, D.~Dohan, E.~Jiang,
  C.~Cai, M.~Terry, Q.~Le, et~al.
\newblock Program synthesis with large language models.
\newblock \emph{arXiv preprint arXiv:2108.07732}, 2021.
\newblock \doi{10.48550/arXiv.2108.07732}.

\bibitem[Bai et~al.(2024)Bai, Lv, Zhang, Lyu, Tang, Huang, Du, Liu, Zeng, Hou,
  Dong, Tang, and Li]{longbench}
Y.~Bai, X.~Lv, J.~Zhang, H.~Lyu, J.~Tang, Z.~Huang, Z.~Du, X.~Liu, A.~Zeng,
  L.~Hou, Y.~Dong, J.~Tang, and J.~Li.
\newblock {L}ong{B}ench: A bilingual, multitask benchmark for long context
  understanding.
\newblock In \emph{Proceedings of the 62nd Annual Meeting of the Association
  for Computational Linguistics (Volume 1: Long Papers)}, pages 3119--3137,
  Bangkok, Thailand, Aug. 2024. Association for Computational Linguistics.
\newblock \doi{10.18653/v1/2024.acl-long.172}.
\newblock URL \url{https://aclanthology.org/2024.acl-long.172}.

\bibitem[Chen et~al.(2021)Chen, Tworek, Jun, Yuan, de~Oliveira~Pinto, Kaplan,
  Edwards, Burda, Joseph, Brockman, Ray, Puri, Krueger, Petrov, Khlaaf, Sastry,
  Mishkin, Chan, Gray, Ryder, Pavlov, Power, Kaiser, Bavarian, Winter, Tillet,
  Such, Cummings, Plappert, Chantzis, Barnes, Herbert-Voss, Guss, Nichol,
  Paino, Tezak, Tang, Babuschkin, Balaji, Jain, Saunders, Hesse, Carr, Leike,
  Achiam, Misra, Morikawa, Radford, Knight, Brundage, Murati, Mayer, Welinder,
  McGrew, Amodei, McCandlish, Sutskever, and Zaremba]{humaneval}
M.~Chen, J.~Tworek, H.~Jun, Q.~Yuan, H.~P. de~Oliveira~Pinto, J.~Kaplan,
  H.~Edwards, Y.~Burda, N.~Joseph, G.~Brockman, A.~Ray, R.~Puri, G.~Krueger,
  M.~Petrov, H.~Khlaaf, G.~Sastry, P.~Mishkin, B.~Chan, S.~Gray, N.~Ryder,
  M.~Pavlov, A.~Power, L.~Kaiser, M.~Bavarian, C.~Winter, P.~Tillet, F.~P.
  Such, D.~Cummings, M.~Plappert, F.~Chantzis, E.~Barnes, A.~Herbert-Voss,
  W.~H. Guss, A.~Nichol, A.~Paino, N.~Tezak, J.~Tang, I.~Babuschkin, S.~Balaji,
  S.~Jain, W.~Saunders, C.~Hesse, A.~N. Carr, J.~Leike, J.~Achiam, V.~Misra,
  E.~Morikawa, A.~Radford, M.~Knight, M.~Brundage, M.~Murati, K.~Mayer,
  P.~Welinder, B.~McGrew, D.~Amodei, S.~McCandlish, I.~Sutskever, and
  W.~Zaremba.
\newblock Evaluating large language models trained on code.
\newblock 2021.

\bibitem[Cobbe et~al.(2021)Cobbe, Kosaraju, Bavarian, Chen, Jun, Kaiser,
  Plappert, Tworek, Hilton, Nakano, et~al.]{gsm8k}
K.~Cobbe, V.~Kosaraju, M.~Bavarian, M.~Chen, H.~Jun, L.~Kaiser, M.~Plappert,
  J.~Tworek, J.~Hilton, R.~Nakano, et~al.
\newblock Training verifiers to solve math word problems.
\newblock \emph{arXiv preprint arXiv:2110.14168}, 2021.
\newblock \doi{10.48550/arXiv.2110.14168}.

\bibitem[Duanmu et~al.(2024)Duanmu, Yuan, Li, Duan, ZHANG, and Lin]{skvq}
H.~Duanmu, Z.~Yuan, X.~Li, J.~Duan, X.~ZHANG, and D.~Lin.
\newblock {SKVQ}: Sliding-window key and value cache quantization for large
  language models.
\newblock In \emph{First Conference on Language Modeling}, 2024.
\newblock URL \url{https://openreview.net/forum?id=nI6JyFSnyV}.

\bibitem[Gao et~al.(2024)Gao, Tow, Abbasi, Biderman, Black, DiPofi, Foster,
  Golding, Hsu, Le~Noac'h, Li, McDonell, Muennighoff, Ociepa, Phang, Reynolds,
  Schoelkopf, Skowron, Sutawika, Tang, Thite, Wang, Wang, and Zou]{lmeval}
L.~Gao, J.~Tow, B.~Abbasi, S.~Biderman, S.~Black, A.~DiPofi, C.~Foster,
  L.~Golding, J.~Hsu, A.~Le~Noac'h, H.~Li, K.~McDonell, N.~Muennighoff,
  C.~Ociepa, J.~Phang, L.~Reynolds, H.~Schoelkopf, A.~Skowron, L.~Sutawika,
  E.~Tang, A.~Thite, B.~Wang, K.~Wang, and A.~Zou.
\newblock A framework for few-shot language model evaluation, 07 2024.
\newblock URL \url{https://zenodo.org/records/12608602}.

\bibitem[Grattafiori et~al.(2024)Grattafiori, Dubey, Jauhri, Pandey, Kadian,
  Al-Dahle, Letman, Mathur, Schelten, Vaughan, Yang, Fan, Goyal, Hartshorn,
  Yang, Mitra, Sravankumar, Korenev, Hinsvark, Rao, Zhang, Rodriguez,
  Gregerson, Spataru, Roziere, Biron, Tang, Chern, Caucheteux, Nayak, Bi,
  Marra, McConnell, Keller, Touret, Wu, Wong, Ferrer, Nikolaidis, Allonsius,
  Song, Pintz, Livshits, Wyatt, Esiobu, Choudhary, Mahajan, Garcia-Olano,
  Perino, Hupkes, Lakomkin, AlBadawy, Lobanova, Dinan, Smith, Radenovic,
  Guzmán, Zhang, Synnaeve, Lee, Anderson, Thattai, Nail, Mialon, Pang,
  Cucurell, Nguyen, Korevaar, Xu, Touvron, Zarov, Ibarra, Kloumann, Misra,
  Evtimov, Zhang, Copet, Lee, Geffert, Vranes, Park, Mahadeokar, Shah, van~der
  Linde, Billock, Hong, Lee, Fu, Chi, Huang, Liu, Wang, Yu, Bitton, Spisak,
  Park, Rocca, Johnstun, Saxe, Jia, Alwala, Prasad, Upasani, Plawiak, Li,
  Heafield, Stone, El-Arini, Iyer, Malik, Chiu, Bhalla, Lakhotia,
  Rantala-Yeary, van~der Maaten, Chen, Tan, Jenkins, Martin, Madaan, Malo,
  Blecher, Landzaat, de~Oliveira, Muzzi, Pasupuleti, Singh, Paluri, Kardas,
  Tsimpoukelli, Oldham, Rita, Pavlova, Kambadur, Lewis, Si, Singh, Hassan,
  Goyal, Torabi, Bashlykov, Bogoychev, Chatterji, Zhang, Duchenne, Çelebi,
  Alrassy, Zhang, Li, Vasic, Weng, Bhargava, Dubal, Krishnan, Koura, Xu, He,
  Dong, Srinivasan, Ganapathy, Calderer, Cabral, Stojnic, Raileanu, Maheswari,
  Girdhar, Patel, Sauvestre, Polidoro, Sumbaly, Taylor, Silva, Hou, Wang,
  Hosseini, Chennabasappa, Singh, Bell, Kim, Edunov, Nie, Narang, Raparthy,
  Shen, Wan, Bhosale, Zhang, Vandenhende, Batra, Whitman, Sootla, Collot,
  Gururangan, Borodinsky, Herman, Fowler, Sheasha, Georgiou, Scialom,
  Speckbacher, Mihaylov, Xiao, Karn, Goswami, Gupta, Ramanathan, Kerkez,
  Gonguet, Do, Vogeti, Albiero, Petrovic, Chu, Xiong, Fu, Meers, Martinet,
  Wang, Wang, Tan, Xia, Xie, Jia, Wang, Goldschlag, Gaur, Babaei, Wen, Song,
  Zhang, Li, Mao, Coudert, Yan, Chen, Papakipos, Singh, Srivastava, Jain,
  Kelsey, Shajnfeld, Gangidi, Victoria, Goldstand, Menon, Sharma, Boesenberg,
  Baevski, Feinstein, Kallet, Sangani, Teo, Yunus, Lupu, Alvarado, Caples, Gu,
  Ho, Poulton, Ryan, Ramchandani, Dong, Franco, Goyal, Saraf, Chowdhury,
  Gabriel, Bharambe, Eisenman, Yazdan, James, Maurer, Leonhardi, Huang, Loyd,
  Paola, Paranjape, Liu, Wu, Ni, Hancock, Wasti, Spence, Stojkovic, Gamido,
  Montalvo, Parker, Burton, Mejia, Liu, Wang, Kim, Zhou, Hu, Chu, Cai, Tindal,
  Feichtenhofer, Gao, Civin, Beaty, Kreymer, Li, Adkins, Xu, Testuggine, David,
  Parikh, Liskovich, Foss, Wang, Le, Holland, Dowling, Jamil, Montgomery,
  Presani, Hahn, Wood, Le, Brinkman, Arcaute, Dunbar, Smothers, Sun, Kreuk,
  Tian, Kokkinos, Ozgenel, Caggioni, Kanayet, Seide, Florez, Schwarz, Badeer,
  Swee, Halpern, Herman, Sizov, Guangyi, Zhang, Lakshminarayanan, Inan,
  Shojanazeri, Zou, Wang, Zha, Habeeb, Rudolph, Suk, Aspegren, Goldman, Zhan,
  Damlaj, Molybog, Tufanov, Leontiadis, Veliche, Gat, Weissman, Geboski, Kohli,
  Lam, Asher, Gaya, Marcus, Tang, Chan, Zhen, Reizenstein, Teboul, Zhong, Jin,
  Yang, Cummings, Carvill, Shepard, McPhie, Torres, Ginsburg, Wang, Wu, U,
  Saxena, Khandelwal, Zand, Matosich, Veeraraghavan, Michelena, Li, Jagadeesh,
  Huang, Chawla, Huang, Chen, Garg, A, Silva, Bell, Zhang, Guo, Yu, Moshkovich,
  Wehrstedt, Khabsa, Avalani, Bhatt, Mankus, Hasson, Lennie, Reso, Groshev,
  Naumov, Lathi, Keneally, Liu, Seltzer, Valko, Restrepo, Patel, Vyatskov,
  Samvelyan, Clark, Macey, Wang, Hermoso, Metanat, Rastegari, Bansal,
  Santhanam, Parks, White, Bawa, Singhal, Egebo, Usunier, Mehta, Laptev, Dong,
  Cheng, Chernoguz, Hart, Salpekar, Kalinli, Kent, Parekh, Saab, Balaji,
  Rittner, Bontrager, Roux, Dollar, Zvyagina, Ratanchandani, Yuvraj, Liang,
  Alao, Rodriguez, Ayub, Murthy, Nayani, Mitra, Parthasarathy, Li, Hogan,
  Battey, Wang, Howes, Rinott, Mehta, Siby, Bondu, Datta, Chugh, Hunt, Dhillon,
  Sidorov, Pan, Mahajan, Verma, Yamamoto, Ramaswamy, Lindsay, Lindsay, Feng,
  Lin, Zha, Patil, Shankar, Zhang, Zhang, Wang, Agarwal, Sajuyigbe, Chintala,
  Max, Chen, Kehoe, Satterfield, Govindaprasad, Gupta, Deng, Cho, Virk,
  Subramanian, Choudhury, Goldman, Remez, Glaser, Best, Koehler, Robinson, Li,
  Zhang, Matthews, Chou, Shaked, Vontimitta, Ajayi, Montanez, Mohan, Kumar,
  Mangla, Ionescu, Poenaru, Mihailescu, Ivanov, Li, Wang, Jiang, Bouaziz,
  Constable, Tang, Wu, Wang, Wu, Gao, Kleinman, Chen, Hu, Jia, Qi, Li, Zhang,
  Zhang, Adi, Nam, Yu, Wang, Zhao, Hao, Qian, Li, He, Rait, DeVito, Rosnbrick,
  Wen, Yang, Zhao, and Ma]{llama3}
A.~Grattafiori, A.~Dubey, A.~Jauhri, A.~Pandey, A.~Kadian, A.~Al-Dahle,
  A.~Letman, A.~Mathur, A.~Schelten, A.~Vaughan, A.~Yang, A.~Fan, A.~Goyal,
  A.~Hartshorn, A.~Yang, A.~Mitra, A.~Sravankumar, A.~Korenev, A.~Hinsvark,
  A.~Rao, A.~Zhang, A.~Rodriguez, A.~Gregerson, A.~Spataru, B.~Roziere,
  B.~Biron, B.~Tang, B.~Chern, C.~Caucheteux, C.~Nayak, C.~Bi, C.~Marra,
  C.~McConnell, C.~Keller, C.~Touret, C.~Wu, C.~Wong, C.~C. Ferrer,
  C.~Nikolaidis, D.~Allonsius, D.~Song, D.~Pintz, D.~Livshits, D.~Wyatt,
  D.~Esiobu, D.~Choudhary, D.~Mahajan, D.~Garcia-Olano, D.~Perino, D.~Hupkes,
  E.~Lakomkin, E.~AlBadawy, E.~Lobanova, E.~Dinan, E.~M. Smith, F.~Radenovic,
  F.~Guzmán, F.~Zhang, G.~Synnaeve, G.~Lee, G.~L. Anderson, G.~Thattai,
  G.~Nail, G.~Mialon, G.~Pang, G.~Cucurell, H.~Nguyen, H.~Korevaar, H.~Xu,
  H.~Touvron, I.~Zarov, I.~A. Ibarra, I.~Kloumann, I.~Misra, I.~Evtimov,
  J.~Zhang, J.~Copet, J.~Lee, J.~Geffert, J.~Vranes, J.~Park, J.~Mahadeokar,
  J.~Shah, J.~van~der Linde, J.~Billock, J.~Hong, J.~Lee, J.~Fu, J.~Chi,
  J.~Huang, J.~Liu, J.~Wang, J.~Yu, J.~Bitton, J.~Spisak, J.~Park, J.~Rocca,
  J.~Johnstun, J.~Saxe, J.~Jia, K.~V. Alwala, K.~Prasad, K.~Upasani,
  K.~Plawiak, K.~Li, K.~Heafield, K.~Stone, K.~El-Arini, K.~Iyer, K.~Malik,
  K.~Chiu, K.~Bhalla, K.~Lakhotia, L.~Rantala-Yeary, L.~van~der Maaten,
  L.~Chen, L.~Tan, L.~Jenkins, L.~Martin, L.~Madaan, L.~Malo, L.~Blecher,
  L.~Landzaat, L.~de~Oliveira, M.~Muzzi, M.~Pasupuleti, M.~Singh, M.~Paluri,
  M.~Kardas, M.~Tsimpoukelli, M.~Oldham, M.~Rita, M.~Pavlova, M.~Kambadur,
  M.~Lewis, M.~Si, M.~K. Singh, M.~Hassan, N.~Goyal, N.~Torabi, N.~Bashlykov,
  N.~Bogoychev, N.~Chatterji, N.~Zhang, O.~Duchenne, O.~Çelebi, P.~Alrassy,
  P.~Zhang, P.~Li, P.~Vasic, P.~Weng, P.~Bhargava, P.~Dubal, P.~Krishnan, P.~S.
  Koura, P.~Xu, Q.~He, Q.~Dong, R.~Srinivasan, R.~Ganapathy, R.~Calderer, R.~S.
  Cabral, R.~Stojnic, R.~Raileanu, R.~Maheswari, R.~Girdhar, R.~Patel,
  R.~Sauvestre, R.~Polidoro, R.~Sumbaly, R.~Taylor, R.~Silva, R.~Hou, R.~Wang,
  S.~Hosseini, S.~Chennabasappa, S.~Singh, S.~Bell, S.~S. Kim, S.~Edunov,
  S.~Nie, S.~Narang, S.~Raparthy, S.~Shen, S.~Wan, S.~Bhosale, S.~Zhang,
  S.~Vandenhende, S.~Batra, S.~Whitman, S.~Sootla, S.~Collot, S.~Gururangan,
  S.~Borodinsky, T.~Herman, T.~Fowler, T.~Sheasha, T.~Georgiou, T.~Scialom,
  T.~Speckbacher, T.~Mihaylov, T.~Xiao, U.~Karn, V.~Goswami, V.~Gupta,
  V.~Ramanathan, V.~Kerkez, V.~Gonguet, V.~Do, V.~Vogeti, V.~Albiero,
  V.~Petrovic, W.~Chu, W.~Xiong, W.~Fu, W.~Meers, X.~Martinet, X.~Wang,
  X.~Wang, X.~E. Tan, X.~Xia, X.~Xie, X.~Jia, X.~Wang, Y.~Goldschlag, Y.~Gaur,
  Y.~Babaei, Y.~Wen, Y.~Song, Y.~Zhang, Y.~Li, Y.~Mao, Z.~D. Coudert, Z.~Yan,
  Z.~Chen, Z.~Papakipos, A.~Singh, A.~Srivastava, A.~Jain, A.~Kelsey,
  A.~Shajnfeld, A.~Gangidi, A.~Victoria, A.~Goldstand, A.~Menon, A.~Sharma,
  A.~Boesenberg, A.~Baevski, A.~Feinstein, A.~Kallet, A.~Sangani, A.~Teo,
  A.~Yunus, A.~Lupu, A.~Alvarado, A.~Caples, A.~Gu, A.~Ho, A.~Poulton, A.~Ryan,
  A.~Ramchandani, A.~Dong, A.~Franco, A.~Goyal, A.~Saraf, A.~Chowdhury,
  A.~Gabriel, A.~Bharambe, A.~Eisenman, A.~Yazdan, B.~James, B.~Maurer,
  B.~Leonhardi, B.~Huang, B.~Loyd, B.~D. Paola, B.~Paranjape, B.~Liu, B.~Wu,
  B.~Ni, B.~Hancock, B.~Wasti, B.~Spence, B.~Stojkovic, B.~Gamido, B.~Montalvo,
  C.~Parker, C.~Burton, C.~Mejia, C.~Liu, C.~Wang, C.~Kim, C.~Zhou, C.~Hu,
  C.-H. Chu, C.~Cai, C.~Tindal, C.~Feichtenhofer, C.~Gao, D.~Civin, D.~Beaty,
  D.~Kreymer, D.~Li, D.~Adkins, D.~Xu, D.~Testuggine, D.~David, D.~Parikh,
  D.~Liskovich, D.~Foss, D.~Wang, D.~Le, D.~Holland, E.~Dowling, E.~Jamil,
  E.~Montgomery, E.~Presani, E.~Hahn, E.~Wood, E.-T. Le, E.~Brinkman,
  E.~Arcaute, E.~Dunbar, E.~Smothers, F.~Sun, F.~Kreuk, F.~Tian, F.~Kokkinos,
  F.~Ozgenel, F.~Caggioni, F.~Kanayet, F.~Seide, G.~M. Florez, G.~Schwarz,
  G.~Badeer, G.~Swee, G.~Halpern, G.~Herman, G.~Sizov, Guangyi, Zhang,
  G.~Lakshminarayanan, H.~Inan, H.~Shojanazeri, H.~Zou, H.~Wang, H.~Zha,
  H.~Habeeb, H.~Rudolph, H.~Suk, H.~Aspegren, H.~Goldman, H.~Zhan, I.~Damlaj,
  I.~Molybog, I.~Tufanov, I.~Leontiadis, I.-E. Veliche, I.~Gat, J.~Weissman,
  J.~Geboski, J.~Kohli, J.~Lam, J.~Asher, J.-B. Gaya, J.~Marcus, J.~Tang,
  J.~Chan, J.~Zhen, J.~Reizenstein, J.~Teboul, J.~Zhong, J.~Jin, J.~Yang,
  J.~Cummings, J.~Carvill, J.~Shepard, J.~McPhie, J.~Torres, J.~Ginsburg,
  J.~Wang, K.~Wu, K.~H. U, K.~Saxena, K.~Khandelwal, K.~Zand, K.~Matosich,
  K.~Veeraraghavan, K.~Michelena, K.~Li, K.~Jagadeesh, K.~Huang, K.~Chawla,
  K.~Huang, L.~Chen, L.~Garg, L.~A, L.~Silva, L.~Bell, L.~Zhang, L.~Guo, L.~Yu,
  L.~Moshkovich, L.~Wehrstedt, M.~Khabsa, M.~Avalani, M.~Bhatt, M.~Mankus,
  M.~Hasson, M.~Lennie, M.~Reso, M.~Groshev, M.~Naumov, M.~Lathi, M.~Keneally,
  M.~Liu, M.~L. Seltzer, M.~Valko, M.~Restrepo, M.~Patel, M.~Vyatskov,
  M.~Samvelyan, M.~Clark, M.~Macey, M.~Wang, M.~J. Hermoso, M.~Metanat,
  M.~Rastegari, M.~Bansal, N.~Santhanam, N.~Parks, N.~White, N.~Bawa,
  N.~Singhal, N.~Egebo, N.~Usunier, N.~Mehta, N.~P. Laptev, N.~Dong, N.~Cheng,
  O.~Chernoguz, O.~Hart, O.~Salpekar, O.~Kalinli, P.~Kent, P.~Parekh, P.~Saab,
  P.~Balaji, P.~Rittner, P.~Bontrager, P.~Roux, P.~Dollar, P.~Zvyagina,
  P.~Ratanchandani, P.~Yuvraj, Q.~Liang, R.~Alao, R.~Rodriguez, R.~Ayub,
  R.~Murthy, R.~Nayani, R.~Mitra, R.~Parthasarathy, R.~Li, R.~Hogan, R.~Battey,
  R.~Wang, R.~Howes, R.~Rinott, S.~Mehta, S.~Siby, S.~J. Bondu, S.~Datta,
  S.~Chugh, S.~Hunt, S.~Dhillon, S.~Sidorov, S.~Pan, S.~Mahajan, S.~Verma,
  S.~Yamamoto, S.~Ramaswamy, S.~Lindsay, S.~Lindsay, S.~Feng, S.~Lin, S.~C.
  Zha, S.~Patil, S.~Shankar, S.~Zhang, S.~Zhang, S.~Wang, S.~Agarwal,
  S.~Sajuyigbe, S.~Chintala, S.~Max, S.~Chen, S.~Kehoe, S.~Satterfield,
  S.~Govindaprasad, S.~Gupta, S.~Deng, S.~Cho, S.~Virk, S.~Subramanian,
  S.~Choudhury, S.~Goldman, T.~Remez, T.~Glaser, T.~Best, T.~Koehler,
  T.~Robinson, T.~Li, T.~Zhang, T.~Matthews, T.~Chou, T.~Shaked, V.~Vontimitta,
  V.~Ajayi, V.~Montanez, V.~Mohan, V.~S. Kumar, V.~Mangla, V.~Ionescu,
  V.~Poenaru, V.~T. Mihailescu, V.~Ivanov, W.~Li, W.~Wang, W.~Jiang,
  W.~Bouaziz, W.~Constable, X.~Tang, X.~Wu, X.~Wang, X.~Wu, X.~Gao,
  Y.~Kleinman, Y.~Chen, Y.~Hu, Y.~Jia, Y.~Qi, Y.~Li, Y.~Zhang, Y.~Zhang,
  Y.~Adi, Y.~Nam, Yu, Wang, Y.~Zhao, Y.~Hao, Y.~Qian, Y.~Li, Y.~He, Z.~Rait,
  Z.~DeVito, Z.~Rosnbrick, Z.~Wen, Z.~Yang, Z.~Zhao, and Z.~Ma.
\newblock The {L}lama 3 herd of models.
\newblock \emph{arXiv preprint arXiv:2407.21783}, 2024.
\newblock \doi{10.48550/arXiv.2407.21783}.

\bibitem[Hariri et~al.(2025)Hariri, Luo, Chen, Zhong, Zhang, Wang, Hu, Han, and
  Chaudhary]{quantizewhatcounts}
M.~Hariri, A.~Luo, W.~Chen, S.~Zhong, T.~Zhang, Q.~Wang, X.~Hu, X.~Han, and
  V.~Chaudhary.
\newblock Quantize what counts: More for keys, less for values.
\newblock \emph{arXiv preprint arXiv:2502.15075}, 2025.
\newblock URL \url{https://arxiv.org/abs/2502.15075}.

\bibitem[Hooper et~al.(2024)Hooper, Kim, Mohammadzadeh, Mahoney, Shao, Keutzer,
  and Gholami]{kvquant}
C.~Hooper, S.~Kim, H.~Mohammadzadeh, M.~W. Mahoney, Y.~S. Shao, K.~Keutzer, and
  A.~Gholami.
\newblock {KVQ}uant: Towards 10 million context length {LLM} inference with
  {KV} cache quantization.
\newblock In A.~Globerson, L.~Mackey, D.~Belgrave, A.~Fan, U.~Paquet,
  J.~Tomczak, and C.~Zhang, editors, \emph{Advances in Neural Information
  Processing Systems}, volume~37, pages 1270--1303. Curran Associates, Inc.,
  2024.
\newblock \doi{10.52202/079017-0040}.

\bibitem[Jiang et~al.(2023)Jiang, Sablayrolles, Mensch, Bamford, Chaplot,
  Casas, Bressand, Lengyel, Lample, Saulnier, et~al.]{mistral}
A.~Q. Jiang, A.~Sablayrolles, A.~Mensch, C.~Bamford, D.~S. Chaplot, D.~Casas,
  F.~Bressand, G.~Lengyel, G.~Lample, L.~Saulnier, et~al.
\newblock Mistral 7{B}. arxiv.
\newblock \emph{arXiv preprint arXiv:2310.06825}, 10:\penalty0 3, 2023.
\newblock \doi{10.48550/arXiv.2310.06825}.

\bibitem[Kang et~al.(2024)Kang, Zhang, Kundu, Jeong, Liu, Krishna, and
  Zhao]{gear}
H.~Kang, Q.~Zhang, S.~Kundu, G.~Jeong, Z.~Liu, T.~Krishna, and T.~Zhao.
\newblock {GEAR}: An efficient error reduction framework for {KV} cache
  compression in {LLM} inference.
\newblock In M.~Rezagholizadeh, P.~Passban, S.~Samiee, V.~Partovi~Nia,
  Y.~Cheng, Y.~Deng, Q.~Liu, and B.~Chen, editors, \emph{Proceedings of The 4th
  NeurIPS Efficient Natural Language and Speech Processing Workshop}, volume
  262 of \emph{Proceedings of Machine Learning Research}, pages 305--321. PMLR,
  14 Dec 2024.
\newblock URL \url{https://proceedings.mlr.press/v262/kang24a.html}.

\bibitem[Lewkowycz et~al.(2022)Lewkowycz, Andreassen, Dohan, Dyer, Michalewski,
  Ramasesh, Slone, Anil, Schlag, Gutman-Solo, Wu, Neyshabur, Gur-Ari, and
  Misra]{minerva}
A.~Lewkowycz, A.~Andreassen, D.~Dohan, E.~Dyer, H.~Michalewski, V.~Ramasesh,
  A.~Slone, C.~Anil, I.~Schlag, T.~Gutman-Solo, Y.~Wu, B.~Neyshabur,
  G.~Gur-Ari, and V.~Misra.
\newblock Solving quantitative reasoning problems with language models.
\newblock In \emph{Proceedings of the 36th International Conference on Neural
  Information Processing Systems}, NIPS '22, Red Hook, NY, USA, 2022. Curran
  Associates Inc.
\newblock ISBN 9781713871088.
\newblock \doi{10.5555/3600270.3600548}.

\bibitem[Liu et~al.(2024{\natexlab{a}})Liu, Liu, Pan, He, Haffari, and
  Zhuang]{minicache}
A.~Liu, J.~Liu, Z.~Pan, Y.~He, G.~Haffari, and B.~Zhuang.
\newblock Mini{C}ache: {KV} cache compression in depth dimension for large
  language models.
\newblock In A.~Globerson, L.~Mackey, D.~Belgrave, A.~Fan, U.~Paquet,
  J.~Tomczak, and C.~Zhang, editors, \emph{Advances in Neural Information
  Processing Systems}, volume~37, pages 139997--140031. Curran Associates,
  Inc., 2024{\natexlab{a}}.
\newblock \doi{10.52202/079017-4443}.
\newblock URL
  \url{https://proceedings.neurips.cc/paper_files/paper/2024/file/fd0705710bf01b88a60a3d479ea341d9-Paper-Conference.pdf}.

\bibitem[Liu et~al.(2024{\natexlab{b}})Liu, Yuan, Jin, Zhong, Xu, Braverman,
  Chen, and Hu]{kivi}
Z.~Liu, J.~Yuan, H.~Jin, S.~Zhong, Z.~Xu, V.~Braverman, B.~Chen, and X.~Hu.
\newblock {KIVI}: A tuning-free asymmetric 2bit quantization for {KV} cache.
\newblock \emph{arXiv preprint arXiv:2402.02750}, 2024{\natexlab{b}}.
\newblock \doi{10.48550/arXiv.2402.02750}.

\bibitem[Pope et~al.(2023)Pope, Douglas, Chowdhery, Devlin, Bradbury, Heek,
  Xiao, Agrawal, and Dean]{kvcache}
R.~Pope, S.~Douglas, A.~Chowdhery, J.~Devlin, J.~Bradbury, J.~Heek, K.~Xiao,
  S.~Agrawal, and J.~Dean.
\newblock Efficiently scaling transformer inference.
\newblock In D.~Song, M.~Carbin, and T.~Chen, editors, \emph{Proceedings of
  Machine Learning and Systems}, volume~5, pages 606--624. Curan, 2023.
\newblock URL
  \url{https://proceedings.mlsys.org/paper_files/paper/2023/file/c4be71ab8d24cdfb45e3d06dbfca2780-Paper-mlsys2023.pdf}.

\bibitem[Sanovar et~al.(2025)Sanovar, Bharadwaj, Amant, R{\"u}hle, and
  Rajmohan]{leanattention}
R.~Sanovar, S.~Bharadwaj, R.~S. Amant, V.~R{\"u}hle, and S.~Rajmohan.
\newblock Lean{A}ttention: Hardware-aware scalable attention mechanism for the
  decode-phase of transformers.
\newblock In \emph{Eighth Conference on Machine Learning and Systems}, 2025.
\newblock URL \url{https://openreview.net/forum?id=KVZDNEoC0Q}.

\bibitem[Su and Yuan(2025)]{attention_sink}
Z.~Su and K.~Yuan.
\newblock {KVS}ink: Understanding and enhancing the preservation of attention
  sinks in {KV} cache quantization for {LLM}s.
\newblock In \emph{Second Conference on Language Modeling}, 2025.
\newblock URL \url{https://openreview.net/forum?id=gIqb6zWZoO}.

\bibitem[Tehrani et~al.(2026)Tehrani, Gao, and Sharma]{turboquant_code}
S.~Tehrani, W.~Gao, and D.~Sharma.
\newblock Turboquant.
\newblock \url{https://github.com/back2matching/turboquant}, 2026.

\bibitem[Touvron et~al.(2023)Touvron, Martin, Stone, Albert, Almahairi, Babaei,
  Bashlykov, Batra, Bhargava, Bhosale, Bikel, Blecher, Ferrer, Chen, Cucurull,
  Esiobu, Fernandes, Fu, Fu, Fuller, Gao, Goswami, Goyal, Hartshorn, Hosseini,
  Hou, Inan, Kardas, Kerkez, Khabsa, Kloumann, Korenev, Koura, Lachaux, Lavril,
  Lee, Liskovich, Lu, Mao, Martinet, Mihaylov, Mishra, Molybog, Nie, Poulton,
  Reizenstein, Rungta, Saladi, Schelten, Silva, Smith, Subramanian, Tan, Tang,
  Taylor, Williams, Kuan, Xu, Yan, Zarov, Zhang, Fan, Kambadur, Narang,
  Rodriguez, Stojnic, Edunov, and Scialom]{llama2}
H.~Touvron, L.~Martin, K.~Stone, P.~Albert, A.~Almahairi, Y.~Babaei,
  N.~Bashlykov, S.~Batra, P.~Bhargava, S.~Bhosale, D.~Bikel, L.~Blecher, C.~C.
  Ferrer, M.~Chen, G.~Cucurull, D.~Esiobu, J.~Fernandes, J.~Fu, W.~Fu,
  B.~Fuller, C.~Gao, V.~Goswami, N.~Goyal, A.~Hartshorn, S.~Hosseini, R.~Hou,
  H.~Inan, M.~Kardas, V.~Kerkez, M.~Khabsa, I.~Kloumann, A.~Korenev, P.~S.
  Koura, M.-A. Lachaux, T.~Lavril, J.~Lee, D.~Liskovich, Y.~Lu, Y.~Mao,
  X.~Martinet, T.~Mihaylov, P.~Mishra, I.~Molybog, Y.~Nie, A.~Poulton,
  J.~Reizenstein, R.~Rungta, K.~Saladi, A.~Schelten, R.~Silva, E.~M. Smith,
  R.~Subramanian, X.~E. Tan, B.~Tang, R.~Taylor, A.~Williams, J.~X. Kuan,
  P.~Xu, Z.~Yan, I.~Zarov, Y.~Zhang, A.~Fan, M.~Kambadur, S.~Narang,
  A.~Rodriguez, R.~Stojnic, S.~Edunov, and T.~Scialom.
\newblock Llama 2: Open foundation and fine-tuned chat models.
\newblock \emph{arXiv preprint arXiv:2307.09288}, 2023.
\newblock \doi{10.48550/arXiv.2307.09288}.

\bibitem[Wang et~al.(2025)Wang, Han, Xu, and Srivastava]{squat}
H.~Wang, L.~Han, K.~Xu, and A.~Srivastava.
\newblock {Sq}uat: Subspace-orthogonal {KV} cache quantization.
\newblock \emph{arXiv preprint arXiv:2503.24358}, 2025.
\newblock \doi{10.48550/arXiv.2503.24358}.

\bibitem[Wolf et~al.(2020)Wolf, Debut, Sanh, Chaumond, Delangue, Moi, Cistac,
  Rault, Louf, Funtowicz, Davison, Shleifer, von Platen, Ma, Jernite, Plu, Xu,
  Scao, Gugger, Drame, Lhoest, and Rush]{huggingface}
T.~Wolf, L.~Debut, V.~Sanh, J.~Chaumond, C.~Delangue, A.~Moi, P.~Cistac,
  T.~Rault, R.~Louf, M.~Funtowicz, J.~Davison, S.~Shleifer, P.~von Platen,
  C.~Ma, Y.~Jernite, J.~Plu, C.~Xu, T.~L. Scao, S.~Gugger, M.~Drame, Q.~Lhoest,
  and A.~M. Rush.
\newblock Transformers: State-of-the-art natural language processing.
\newblock In \emph{Proceedings of the 2020 Conference on Empirical Methods in
  Natural Language Processing: System Demonstrations}, pages 38--45, Online,
  Oct. 2020. Association for Computational Linguistics.
\newblock URL \url{https://www.aclweb.org/anthology/2020.emnlp-demos.6}.

\bibitem[Xiao et~al.(2024)Xiao, Tian, Chen, Han, and Lewis]{streamingllm}
G.~Xiao, Y.~Tian, B.~Chen, S.~Han, and M.~Lewis.
\newblock Efficient streaming language models with attention sinks.
\newblock In \emph{The Twelfth International Conference on Learning
  Representations}, 2024.
\newblock URL \url{https://openreview.net/forum?id=NG7sS51zVF}.

\bibitem[Zandieh et~al.(2026)Zandieh, Daliri, Hadian, and Mirrokni]{turboquant}
A.~Zandieh, M.~Daliri, M.~Hadian, and V.~Mirrokni.
\newblock Turboquant: Online vector quantization with near-optimal distortion
  rate.
\newblock In \emph{The Fourteenth International Conference on Learning
  Representations}, 2026.
\newblock URL \url{https://openreview.net/forum?id=tO3ASKZlok}.

\end{thebibliography}
